\documentclass[conference]{IEEEtran}
\IEEEoverridecommandlockouts

\usepackage{microtype}
\usepackage{graphicx}
\usepackage{caption}
\usepackage{subcaption}
\usepackage{booktabs} % for professional tables
\usepackage{tabularx}
\usepackage{multicol}
\usepackage{empheq}
\usepackage{amsmath}
\usepackage{todonotes}
\usepackage{amsmath,graphicx,multicol}
\usepackage{amsfonts}
\usepackage{color}
\usepackage{bbm}
\usepackage{cite}
\usepackage{amssymb}
\usepackage{xspace} 
\usepackage{multirow}
\usepackage{dirtytalk}
\usepackage{algorithm}
\usepackage{algpseudocode}
\usepackage{tikz}
\usepackage[
  separate-uncertainty = true,
  multi-part-units = repeat
]{siunitx}
\usepackage{adjustbox}

\usepackage{fancyhdr}
\newcommand{\approach}{{\sc APT}\xspace}

\newcommand{\quality}{{\sc RankIQA}\xspace}

\newcommand{\etc}{\emph{etc.}\xspace}

\newcommand{\ie}{\emph{i.e.,}\xspace}
\newcommand{\eg}{\emph{e.g.,}\xspace}

\newcommand{\comment}[1]{{\em \color{red} {#1}}}
\newcommand{\response}[1]{{\em \color{blue} {#1}}}
\newcommand{\cut}[1]{}

\usepackage{hyperref}

\newcommand{\figref}[1]{Figure \ref{#1}}
\newcommand{\secref}[1]{Section \ref{#1}}
\newcommand{\tabref}[1]{Table \ref{#1}}

\begin{document}

%\title{perceptual-quality-RL}
\title{APT: \underline{A}daptive \underline{P}erceptual quality based camera \underline{T}uning using reinforcement learning
{\footnotesize \textsuperscript{}}
\thanks{\IEEEauthorrefmark{2} Work mostly done as an intern at NEC Laboratories America, Inc.}
}
%\author{chak }
%\date{August 2022}

%\author{\IEEEauthorblockN{Sibendu Paul,} \and
%\IEEEauthorblockN{Kunal Rao,} \and 
%\IEEEauthorblockN{Giuseppe Coviello,} \and 
%\IEEEauthorblockN{Murugan Sankaradas,} \and 
%\IEEEauthorblockN{Oliver Po,} \and 
%\IEEEauthorblockN{Y. Charlie Hu,} \and
%\IEEEauthorblockN{Srimat T. Chakradhar}
%}

\author{
    \IEEEauthorblockN{Sibendu Paul\IEEEauthorrefmark{2}, Kunal Rao\IEEEauthorrefmark{1}, Giuseppe Coviello\IEEEauthorrefmark{1}, Murugan Sankaradas\IEEEauthorrefmark{1}, Oliver Po\IEEEauthorrefmark{1}, \\ Y. Charlie Hu\IEEEauthorrefmark{3} and Srimat Chakradhar\IEEEauthorrefmark{1}}
    \IEEEauthorblockA{\IEEEauthorrefmark{1}\{sipaul, kunal, giuseppe.coviello, murugs, oliver, chak\}@nec-labs.com}
    \IEEEauthorblockA{\IEEEauthorrefmark{2}\IEEEauthorrefmark{3}\{paul90, ychu\}@purdue.edu}
    \IEEEauthorblockA{\IEEEauthorrefmark{1}\textit{NEC Laboratories America, Inc.} - Princeton, NJ, USA}
    \IEEEauthorblockA{\IEEEauthorrefmark{2}\IEEEauthorrefmark{3}\textit{Purdue University} - West Lafayette, PA, USA}
}

\maketitle

\vspace{-0.05in}
\begin{abstract}
Cameras are increasingly being deployed in cities, enterprises and roads world-wide
to enable many applications in public safety,
intelligent transportation, retail, healthcare and manufacturing.
Often, after initial deployment of the cameras, the environmental conditions and the scenes around these cameras change, and our experiments show that these changes can adversely impact the accuracy of insights from video analytics.
%Camera vendors do expose several camera parameters that can be 
%tuning the quality of the video produced by the camera. 
This is because the camera parameter settings, though optimal at deployment time, are not the best settings for good-quality video capture as the environmental conditions and scenes around a camera change during operation. Capturing poor-quality video adversely affects the accuracy of analytics. 
To mitigate the loss in accuracy of insights, we propose a novel, reinforcement-learning based system \approach that dynamically, and remotely (over 5G networks),  tunes the camera parameters, 
% By dynamically changing the camera parameters,  we 
to ensure a high-quality video capture, which mitigates any loss in accuracy of video analytics. 
As a result, such tuning restores the accuracy of insights  when environmental conditions or scene content change. 
%The proposed system \approach ensures that good camera settings are %in place at all times during operation of the application or service. 
\approach uses reinforcement learning, with no-reference perceptual quality
estimation as the reward function.
We conducted
extensive real-world experiments, where we simultaneously deployed two cameras side-by-side overlooking an enterprise parking lot (one camera only has manufacturer-suggested default setting, while the other camera is dynamically tuned by
\approach during operation). Our experiments demonstrated 
that due to dynamic tuning by \approach, the analytics
insights are consistently better at all times of the day: 
the accuracy of object detection video analytics application
was improved on average by $\sim$ 42\%. Since our reward function is independent of any analytics task, \approach can be readily used for different video analytics tasks.

\end{abstract}

\vspace{-0.05in}
\section{Introduction}
\label{sec:intro}

The number of IoT sensors, especially video cameras deployed around
the world have proliferated tremendously. It is estimated that their
number will continue to grow further, thanks to advances in computer vision,
machine learning, etc. and infrastructure support through 5G, edge
computing, cloud computing, etc.
%With the rise in 5G, edge computing and advances in computer vision, machine learning, etc. several new applications are emerging.  
These video cameras are being used for a variety of applications including
video surveillance, intelligent transportation, healthcare, retail,
entertainment, safety and security, and home and building automation.
The global video surveillance camera market, which was valued at
US \$28.02 billion in 2021, is estimated to reach US \$45.54 billion in
2027. Furthermore, the volume, which was 214.3 million units in 2021,
is estimated to reach 524.75 million units in 2027~\cite{report-linker}. 
As the video camera market grows, the 
% automatic
video analytics market also grows with it. The global video analytics market
is estimated to grow from \$5 billion in 2020 to \$21 billion by 2027,
at a CAGR of 22.70\% ~\cite{allied-market-research}.
%The global surveillance camera market was valued at US \$28 billion in 2021 and the volume was $\sim$ 210 million units. The valuation is estimated to reach US \$45.5 billion and the volume to reach 525 million units in 2027. The global video analytics market is estimated to grow from \$5 billion in 2020 to \$21 billion by 2027, at a CAGR of 22.70\% ~\cite{allied-market-research}. 
%is going to grow to \textcolor{blue}{add number} by the year  \textcolor{blue}{year [add citation]}. 

\begin{figure}[tp]
    \centering
    \includegraphics[width=0.95\linewidth]{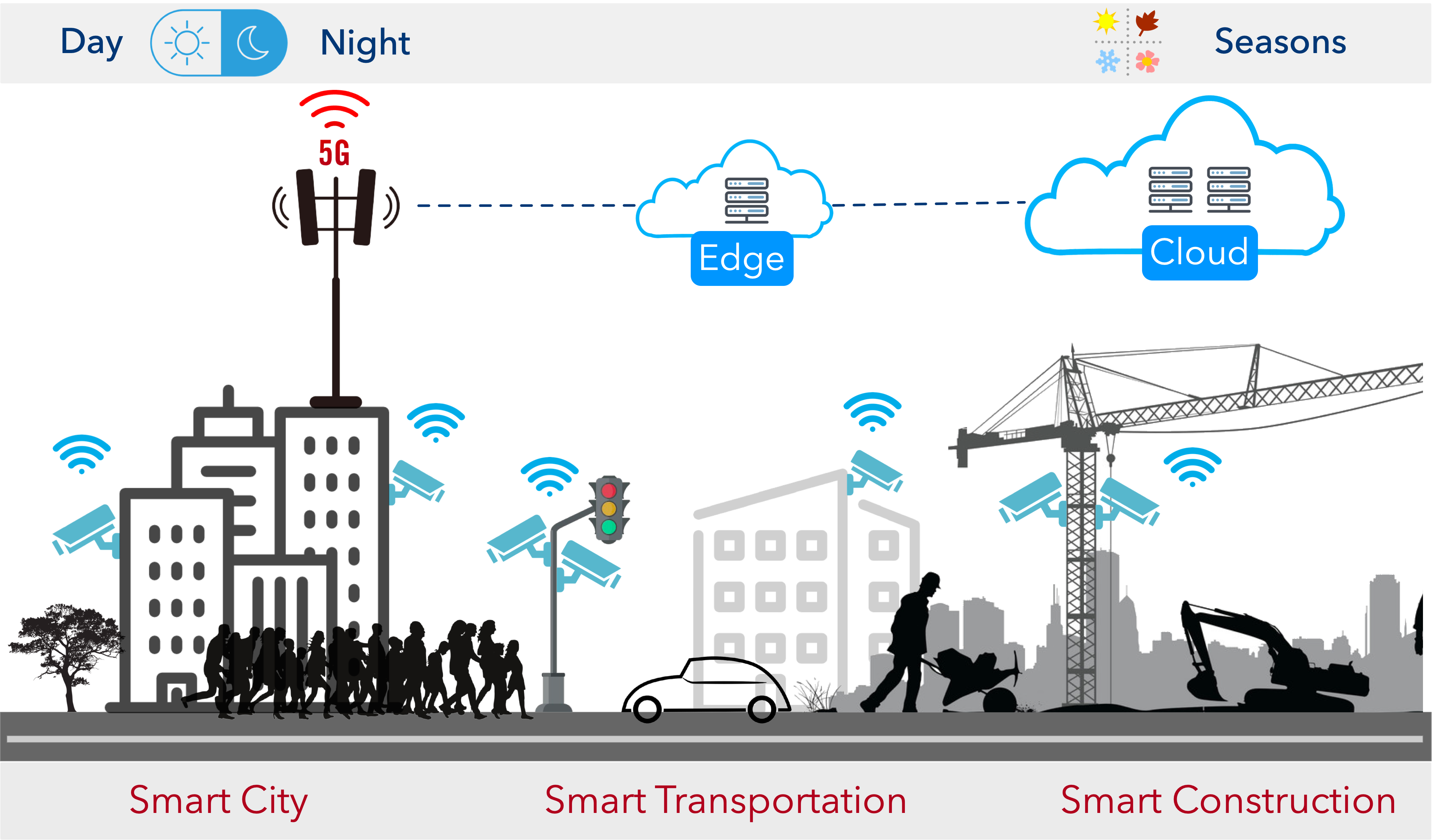}
    %\vspace{-0.2in}
    \caption{City-scale video analytics.}
    \label{fig:intro}
    \vspace{-0.1in}
\end{figure}

A city-scale deployment of IoT cameras and video analytics being
performed on those cameras is shown in \figref{fig:intro}. Here, the
video feed from cameras is streamed over 5G and the analytics is being
performed in the edge/cloud infrastructure. Based on the results of
the analytics, insights are generated and appropriate actions are
taken. As the environmental conditions around the cameras (e.g. day or
night, seasonal variations) and the scene in front of the camera
(e.g. number of people/cars) change, the quality of video feed
produced by the cameras also changes. This is due to the manner in which
cameras capture, process, encode and transmit video frames before they
are delivered to Analytics Units (AUs) in a Video Analytics Pipeline
(VAP).

%, which in turn affects the analytics accuracy. 

%In this paper, we show that when there is variation in the environmental conditions or in the scene in front of the camera, the accuracy of video analytics application will be impacted and may even degrade. 
%In this paper, we show that variation in environmental conditions or in the scene in front of the camera, impacts accuracy of video analytics and it may even degrade. 
In this paper, we show that the accuracy of video analytics application is impacted by variation in environmental conditions or in the scene in front of the camera, and it may even degrade. 
One of the reasons for this degradation is the poor quality
of frames being delivered to the AUs. Camera vendors often expose a
large number of camera parameter settings to end users so that they
can tune them according to their deployment location. These camera
parameters play a significant role in the quality of frames being
produced by the camera and delivered to the AUs. 
We show that if we
adjust these camera parameter values, we can improve the quality
of frames and thereby mitigate the loss in analytics accuracy due to
changes in environment or video content. This adjustment in camera
parameter values however is non-trivial. We observe that the
desirable adjustment values vary according to the condition and is very specific
to the deployment location. There is not any single adjustment setting
that works across all conditions and across all deployment
locations. Therefore, we need a dynamic adjustment technique that
automatically adapts to the changing conditions at the specific
deployment location.

%so as to improve the quality of frames being deliverd to the AUs.

%is different for different conditions and hence is not a one-time effort, rather a continuous 

%If the frames are of poor quality, then the analytics suffers. 

%To mitigate this loss in accuracy due to environmental condition change
To mitigate the loss in analytics accuracy, we propose to adaptively tune camera parameters in real time using reinforcement
learning. In particular, we propose to develop a system that dynamically
tunes four camera parameters,
i.e., brightness, color, contrast and sharpness, which directly affect
the quality of image produced by the video camera. Such dynamic tuning
of camera parameters happens remotely over 5G and leads to better
quality of the video feed, which directly
helps in improving the analytics accuracy.

%For reinforcement learning technique, we need to define the state, action and reward for the agent to learn and act appropriately for different conditions. We define

In designing reinforcement learning, for the agent to learn and adapt to
the changes in conditions, we use perceptual no-reference quality
estimator as the reward function. We show via experiments
that such a reward function works well in adjusting camera settings so
that the loss in analytics accuracy is mitigated. This technique is
independent of the video analytics being performed, and therefore is easy
to design and deploy in real-world settings.

There are different methods to calculate the perceptual quality, but
testing and comparing them to check which one works the best for
real-world setting is not straight forward. First, if we test these
methods one by one, then it is impossible to repeat exact same
environment and video content changes, if the video analytics system (VAS)
is deployed in the wild. Such
a setup will not lead to ``apples to apples'' comparison. Second, if we
test these methods at the same time, then it is not practical to
simultaneously deploy as many cameras as the number of methods to
calculate perceptual quality for each method on each camera at the
same time. These challenges lead us to consider a mock
 experimental setup, which allows us to
repeat the environment and content changes in a controlled setting,
and objectively test and compare different perceptual quality
estimators one-by-one.

In summary, our key contributions are as follows:

\begin{itemize}
    %\item We empirically show that fixed set of values for camera parameters (\ie manufacturer-provided default values) does not lead to consistent high-quality video analytics accuracy as environmental condition and content change, rather adapting the built-in camera parameters helps in enhancing the analytics accuracy. 
    \item We empirically show that changes in environmental conditions and video content can have adverse effect on video analytics accuracy, and this loss in accuracy can be mitigated by dynamically tuning camera settings.
    %\item We propose a Reinforcement Learning (RL) based system \approach, that automatically and adaptively learns the best settings for cameras deployed in the field in reaction to change in environment and content so as to yield consistent high-quality analytical insights.
    \item We propose novel Reinforcement Learning (RL) based system called \approach, which automatically and adaptively tunes camera parameters remotely over 5G, in order to produce good quality video feed, which directly helps in improving analytics insights
    %\item We use CNN-based state-of-the-art perceptual quality estimator (\ie \emph{RankIQA}) as the reward function to make RL based camera-parameter-tuning design feasible in absence of ground-truth.
    \item We use {CNN}-based state-of-the-art perceptual quality estimator (\ie \emph{RankIQA}) as the reward function in RL, thus making \approach\ design independent of the analytics being performed and feasible in absence of ground-truth.
    \item Our adaptive camera-parameter-tuning results in consistent analytics accuracy improvement through different time segments of the day and achieves an average improvement of $\sim$ 42\% when compared to the accuracy observed under fixed, manufacturer-provided default setting.
    %when deployed to monitor an enterprise parking lot. 
    %We observe an average improvement of $\sim$ 42\% more true-positive object detections from the camera stream tuned by \approach.
    %than the default camera stream. 
\end{itemize}

The rest of the paper is organized as follows. We discuss related works in \secref{sec:related}. \secref{sec:motivation} presents the negative impact of environmental condition and content changes on analytics accuracy and how it can be possibly mitigated by adaptively tuning built-in camera parameters. The design challenges of such adaptive camera-parameter-tuning system and final \approach design are shown in \secref{sec:challenge} and \secref{sec:design}, respectively. Extensive evaluation of \approach on 3D mock-up scene as well as real-world deployment is discussed in \secref{sec:eval}.
%. We present several recent efforts on analytics accuracy improvement through tuning several parameters in \secref{sec:related}. 
Finally, in \secref{sec:future} we show how to possibly extend \approach design in the future and conclude in \secref{sec:conc}.

%This is because, if we deploy and test them one by one in a real-world setup, then it is impossible to ensure exact same environment and video content 

%This is because, we need to deploy, test and compare them under identical conditions, and it is not practical to simultaneously deploy as many cameras as the number of ways to calculate perceptual quality. As an alternative method to objectively compare them, we setup a mock scene where we can control 

%to test different techniques one by one in a controlled setting where we can mimic/repeat exact same conditions and observe how each technique performs and then pick the best one for real-world deployment. 

%This mock setup allows us to repeat our experiments with different quality estimators  

%video cameras expose a number of different settings and when we change these settings 

\vspace{-0.05in}
\section{Related Works}
\label{sec:related}
Several recent proposals have investigated the tuning of parameters of
vision algorithms to improve computing resource usage of video
analytics pipelines based on input video
content. Chameleon~\cite{jiang2018chameleon},
Videostorm~\cite{201465}, and AWStream~\cite{zhang2018awstream} tune
the after-capture video stream parameters like frame sampling rate,
frame resolution, type of detector to ensure efficient resource usage
while processing video analytics queries at scale. However, they do
not address directly tuning of camera parameters to enhance video
analytics accuracy.  
%Wang et. al.~\cite{234849} have proposed to send the video stream in low resolution from the edge to the cloud over a wide-area network to reduce the bandwidth requirement, and then recover the high-resolution frames from the low-resolution stream via a super-resolution procedure tailored for the actual analytics tasks. A recent proposal ~\cite{du2020server} also suggests compressing regions of little or no interest more heavily while compressing regions of interest with a lower compression rate to better use the scarce network bandwidth between the sensor and the analytics server. Again, none of these approaches consider adaptive tuning of camera parameters to adapt to changing environmental conditions.

A recent work~\cite{jang2018application} also reports the impact of
environmental condition changes on video analytics accuracy but it
adapts to such changes by using different AUs depending on
specific environmental condition, while keeping the camera settings the same. Since environmental changes can take
place due to change of the sun's movement throughout the day,
different weather conditions (\eg rain, fog and snow), as well as for
different deployment sites (\eg parking lot, shopping mall, airport),
 it is infeasible to develop a separate AU specific to each
environment. Impact of environmental changes on AU accuracy is also shown in ~\cite{video-analytics-accuracy-fluctuation}, but they address it by re-training the AU using transfer learning. In contrast, \approach\ takes a different approach where the AU is kept the same, but camera settings are dynamically tuned in reaction to changes in environmental conditions.

%In contrast, \approach takes a different approach and dynamically adapts the camera settings to different environmental condition changes while keeping the downstream AU fixed.
Several recent works like AMS ~\cite{AMS} and Ekya ~\cite{Ekya} aim to improve video analytics accuracy by periodically re-training AI/ML models so that they work well for the specific deployment conditions. This technique however, requires additional computational resources and it does not quickly adapt to the changes in the environment or video content. \approach, on the other hand does not rely on continuous re-training, rather it improves video analytics accuracy by dynamically tuning configurable camera parameters, thereby quickly reacting to the changes in environmental conditions or video content.

There is a considerable body of work to configure image signal
processing pipeline (ISP) in cameras to improve camera capture
quality.  For example, VisionISP~\cite{wu2019visionisp} modifies the
ISP pipeline to reduce the size of final image output by reducing the
bit-depth and resolution. Others have proposed custom optimizations of
the ISP for specific computer-vision tasks~\cite{heide2014flexisp,
  wu2019visionisp, schwartz2018deepisp, zhang2018ffdnet, liu2020joint,
  diamond2021dirty, nishimura2018automatic}. However, careful
re-design or optimization of ISP module for specific vision tasks is
time consuming. In our proposed approach \approach, we do not modify
the ISP pipeline, rather we focus on dynamic tuning of configurable
camera parameters to consistently produce high-quality video output, which enhances the quality of insights from analytics tasks.

\vspace{-0.05in}
\section{Motivation}
\label{sec:motivation}

%\subsection{Impact of Environment change on AU accuracy}
%\label{subsec:environ}
\if 0
\begin{figure}[tb]
\begin{subfigure}[]{0.48\linewidth}
\centering
    \includegraphics[height=1.25 in]{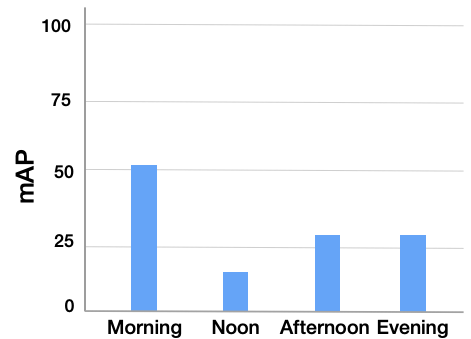}
    \caption{Face detection}
    \label{fig:daylong_face_det}
\end{subfigure}%
\begin{subfigure}[]{0.48\linewidth}
\centering
    \includegraphics[height=1.25 in]{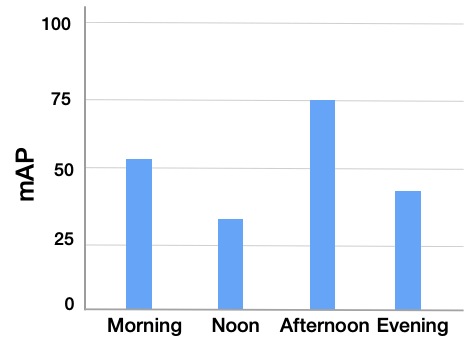}
    \caption{Person detection}
    \label{fig:daylong_person_det}
\end{subfigure}
    \vspace{-0.1in}
 \caption{AU accuracy variation in a day under the default camera setting \textcolor{red}{need some other figure}.}
  \label{fig:face_person_env}
 \vspace{-0.1in}
\end{figure}
\fi

%\subsection{Impact of Camera Settings on AU Accuracy}
%\label{subsec:impactofsettings}

In this section, we show that  environmental conditions and video content variation can adversely impact analytics accuracy and this loss in accuracy can be mitigated by adjusting camera parameter values. To illustrate the impact of environment and content variation on AU accuracy, we
consider four popular parameters that are exposed by almost all
cameras: {\em brightness, contrast, color-saturation} (also known as
colorfulness), and {\em sharpness}.  {We focused on these four
  parameters because they are widely available in both PTZ and non-PTZ
  cameras 
  %without internal automatic tuning 
  and these parameters are more
  challenging to tune due to their large range of parameter values (for
  example between 1 and 100)}. 

\textbf{Methodology:} Analyzing the impact of camera settings on video
analytics poses a significant challenge: it requires applying
different camera parameter settings to the same input scene and
measuring the difference in the resulting accuracy of insights from an AU. The
straight-forward approach is to use multiple cameras with different
camera parameter settings to record the same input scene. However,
such an approach is impractical as there are thousands of different
combinations of even just the four camera parameters we consider.  
To overcome the challenge, we proceed with two workarounds.  
First, we will show the impact of camera settings on a stationary scene with a real camera.  
Second, we apply post-capture image transformation on pre-recorded video snippets from public datasets to analyze the equivalent impact of different camera settings on those video snippets, \ie groups of frames.

\subsection{Impact of environment variation on AU accuracy}  To study the impact of environmental changes on AU performance, we simulate {DAY} and {NIGHT}
conditions in our lab and evaluate the performance of the most
accurate face-recognition AU (Neoface-v3~\cite{NIST}\footnote{ This
  face-recognition AU is ranked first in the world in the most recent
  face-recognition technology benchmarking by NIST.}.  We use two
sources of light and keep one of them always {ON}, while the
other light is manually turned {ON} or {OFF} to emulate DAY and NIGHT
conditions, respectively.  \cut{ When both light sources are
  \emph{ON}, it is bright, thereby simulating the {DAY} condition;
  when one of them is \emph{OFF}, it is low-light, thereby simulating
  the {NIGHT} condition.  }

\begin{figure}[tb]
\begin{subfigure}[]{0.5\linewidth}
\centering
    \includegraphics[width=1.6 in]{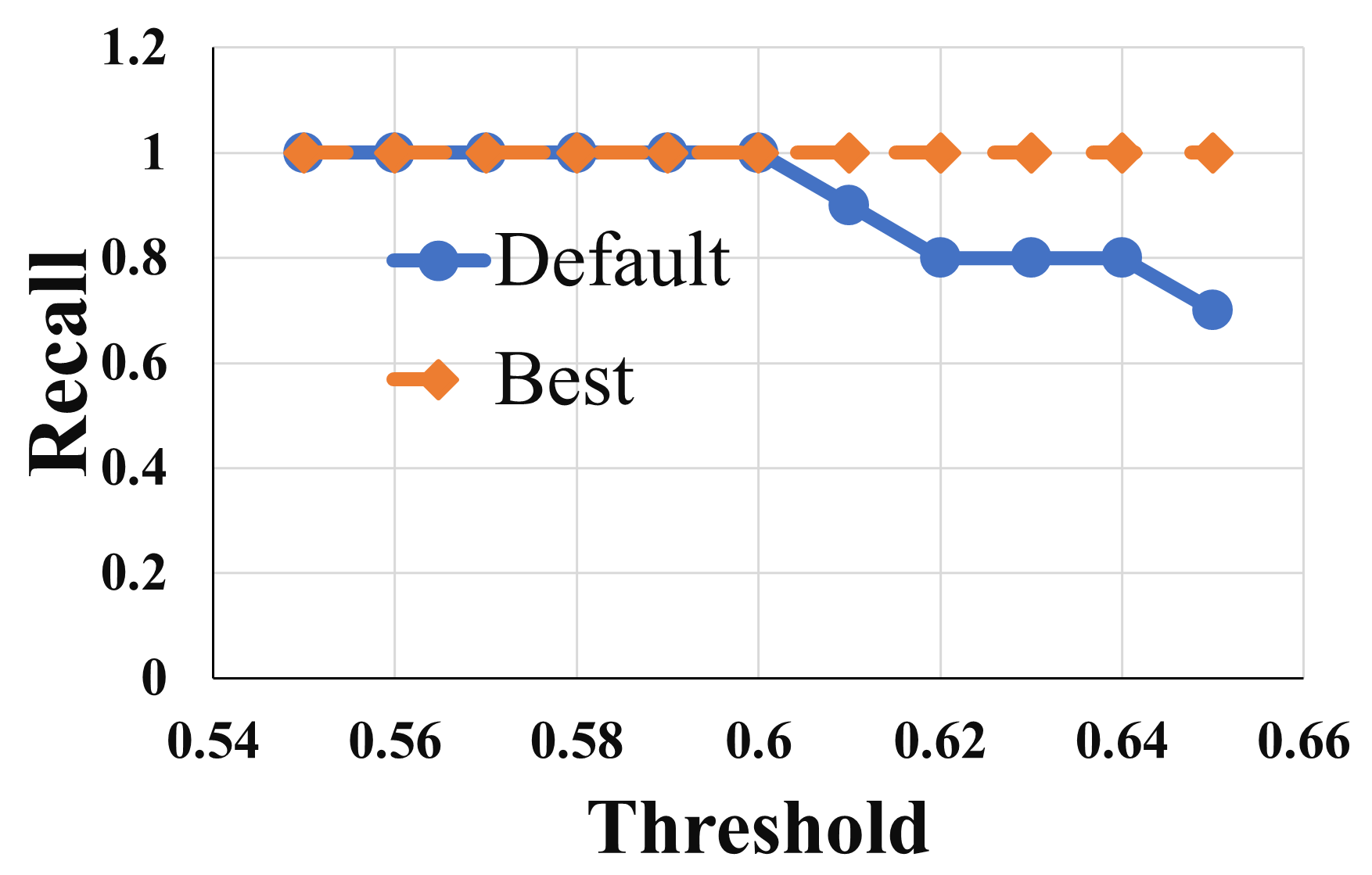}
    \caption{DAY}
    \label{fig:recall_day}
\end{subfigure}%
\begin{subfigure}[]{0.5\linewidth}
\centering
    \includegraphics[width=1.6 in]{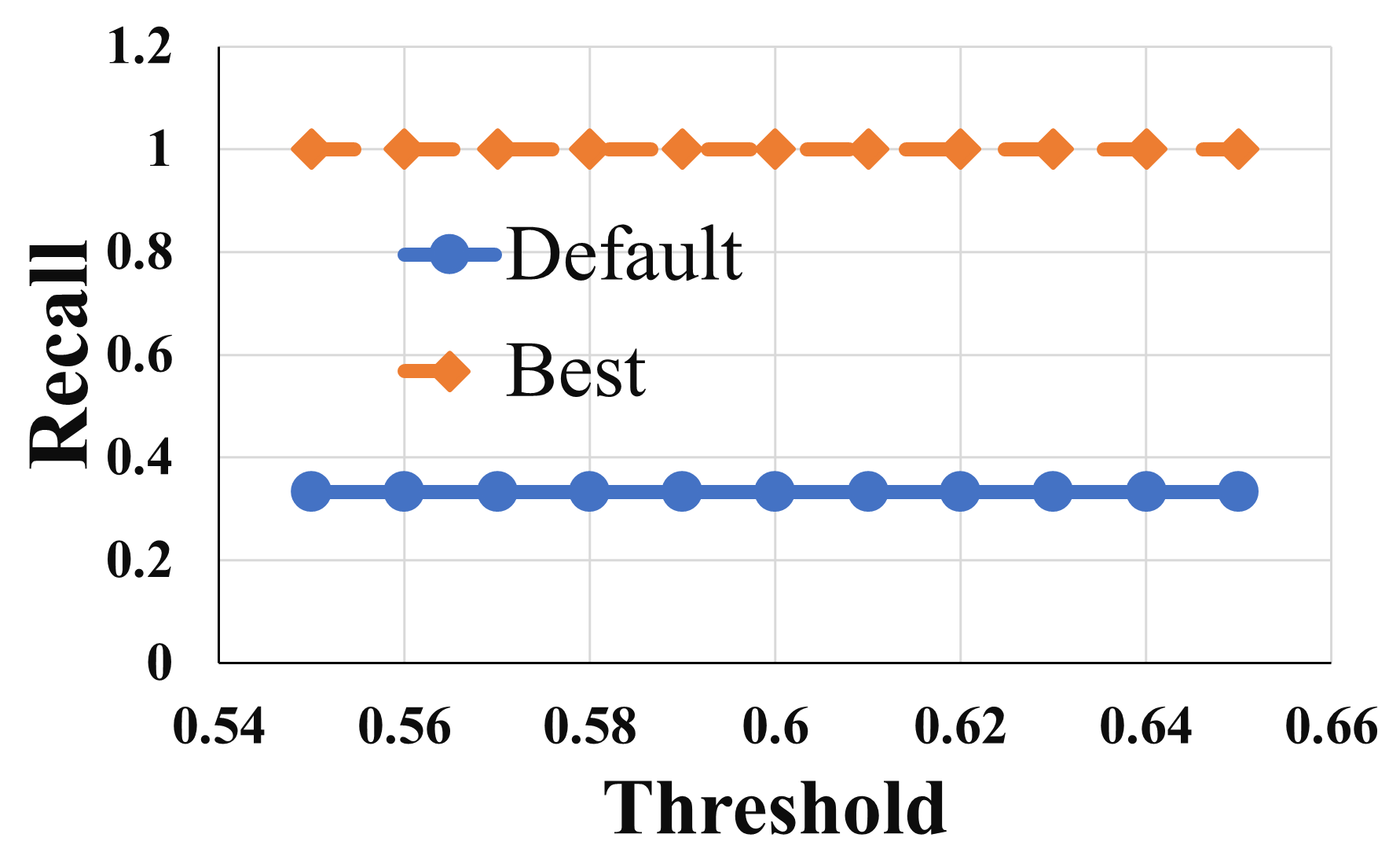}
    \caption{NIGHT}
    \label{fig:recall_night}
\end{subfigure}
 \caption{Parameter tuning impact for Face-recognition AU.}
  \label{fig:recall-phy}
 \vspace{-0.2in}
\end{figure}

We place face cutouts of 12 unique individuals in front of the camera and first run the face recognition pipeline with the input scene captured under ``Default" camera setting (\ie the default
values provided by the manufacturer) and also for different face matching thresholds. Since this face-recognition AU 
has high precision despite environment changes, we focus on measuring Recall, \ie true-positive rate. ~\figref{fig:recall_day} shows the Recall for the {DAY} condition
for various thresholds and ~\figref{fig:recall_night} shows the Recall
for the NIGHT condition for various thresholds. We see that under the ``Default''
settings, the Recall for the {DAY} condition goes down at higher thresholds,
indicating that some faces were not recognized, whereas for
the {NIGHT} condition, the Recall remains constant at a low value for
all thresholds, indicating that some faces were not being recognized
regardless of the face matching thresholds. Thus, the performance of face-recognition AU (\ie recall vs matching threshold) under ``Default" camera setting varies for different environment while capturing the same static scene.
Next, we compare AU results under the ``Default" camera settings,
and ``Best'' settings for the four camera parameters.
To find the ``Best'' settings, we 
change the four camera parameters using the VAPIX API~\cite{vapix}
provided by the camera vendor to find the setting that gives the
highest Recall value. 
Specifically, we vary each parameter from 0 to 100 in steps of
10 and capture the frame for each camera setting. This gives us
$\approx$14.6K (11$^4$) frames 
for each condition.
Changing one camera setting 
through the VAPIX API takes about 200ms,
and in total it took about \emph{7 hours} to capture and process the
frames for each condition. 

In contrast, when we changed the camera parameters for both conditions to the
``Best" settings, the AU achieves the highest Recall (100\%), 
confirming that all the faces are correctly recognized, also shown in~\figref{fig:recall-phy}. These results
show that it is indeed possible to improve AU accuracy by adjusting
the four camera parameters.

%define virtual knobs and impacts
\subsection{Impact of video content variation on AU accuracy}
We study the impact of video content variation on AU accuracy by using pre-recorded videos with different video content. The pre-recorded videos from public datasets  are already
  captured under certain camera parameter settings, and hence we do not have the
  opportunity to change the real camera parameters and observe their
  impact. As an approximation, we apply different
  values of brightness, contrast, color-saturation and sharpness to these
pre-recorded videos using
  several image transformation algorithms in the Python Imaging
  Library ({PIL})~\cite{pil}, and then observe the impact of such
  transformation on accuracy of AU insights.

\begin{figure}[tb]
\begin{subfigure}[]{0.5\linewidth}
\centering
    \includegraphics[height=1.3 in]{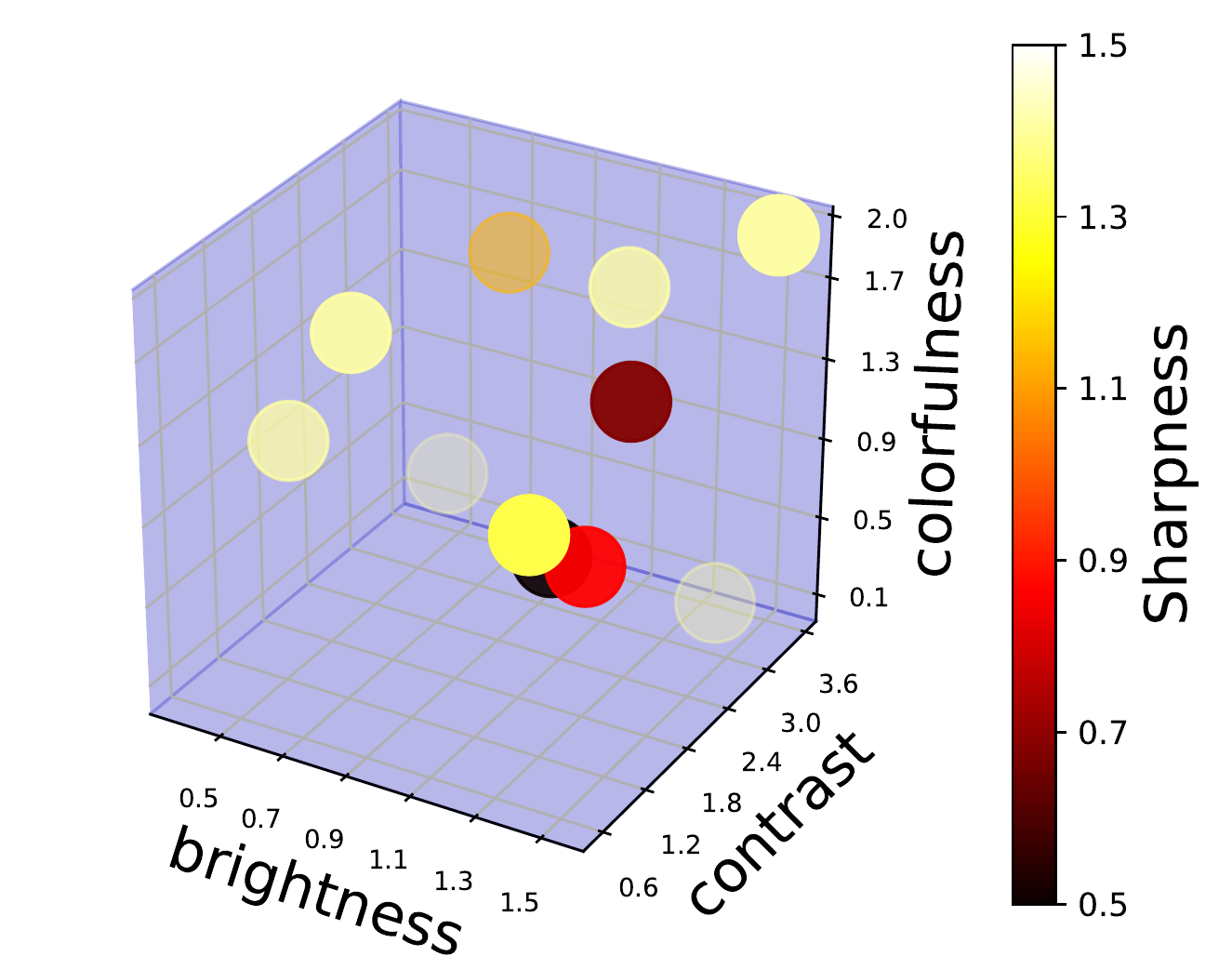}
    \caption{Face detection}
    \label{fig:hmdb_face}
\end{subfigure}%
\begin{subfigure}[]{0.5\linewidth}
\centering
    \includegraphics[height=1.3 in]{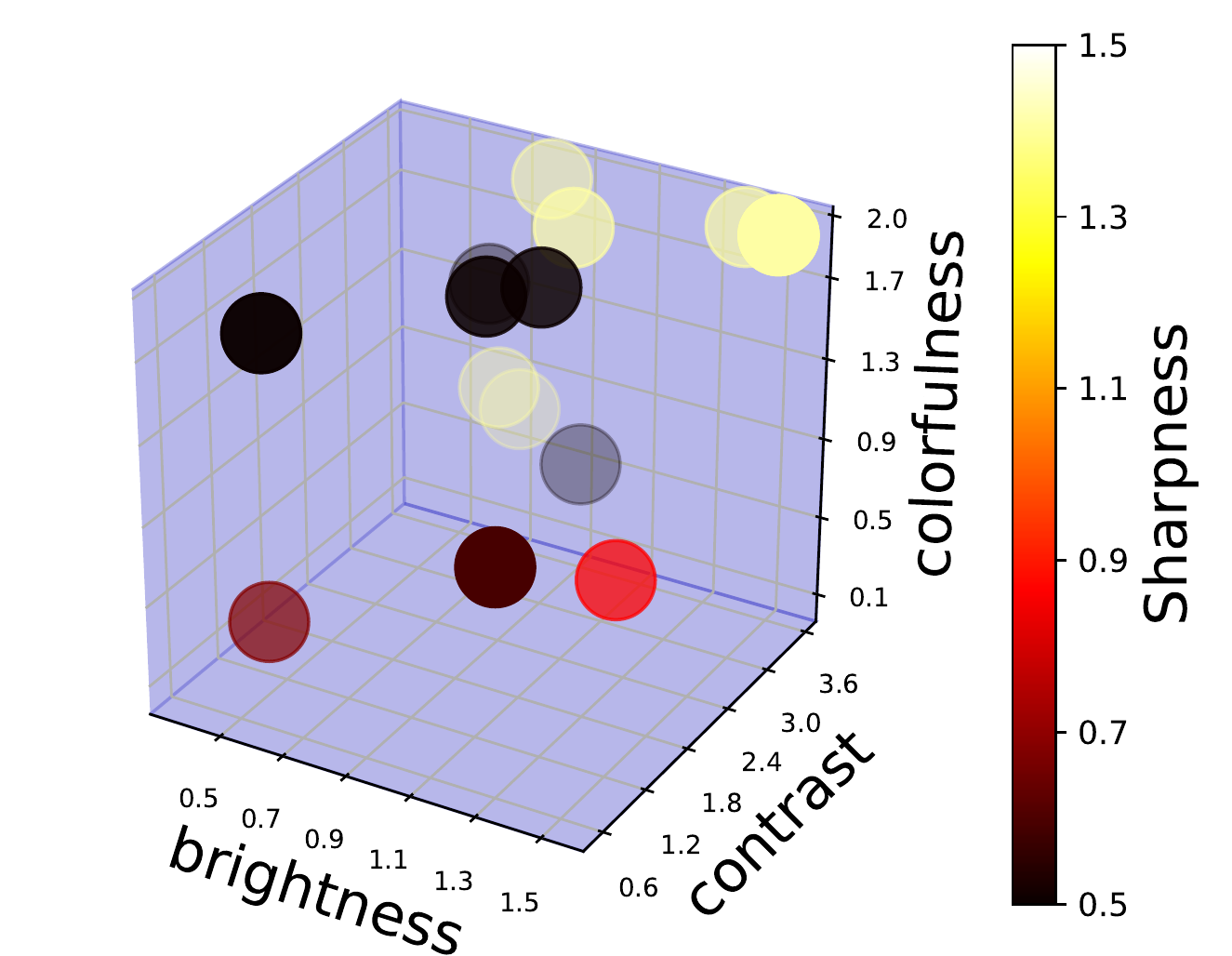}
    \caption{Person detection}
    \label{fig:hmdb51_person}
\end{subfigure}
\caption{Distribution of best \emph{transformation tuple} for two AUs on \emph{HMDB} video snippets.}
\label{fig:hmdb51}
 \vspace{-0.2in}
\end{figure}

\begin{figure}[tb]
\begin{subfigure}[]{0.5\linewidth}
\centering
    \includegraphics[height=1.3 in]{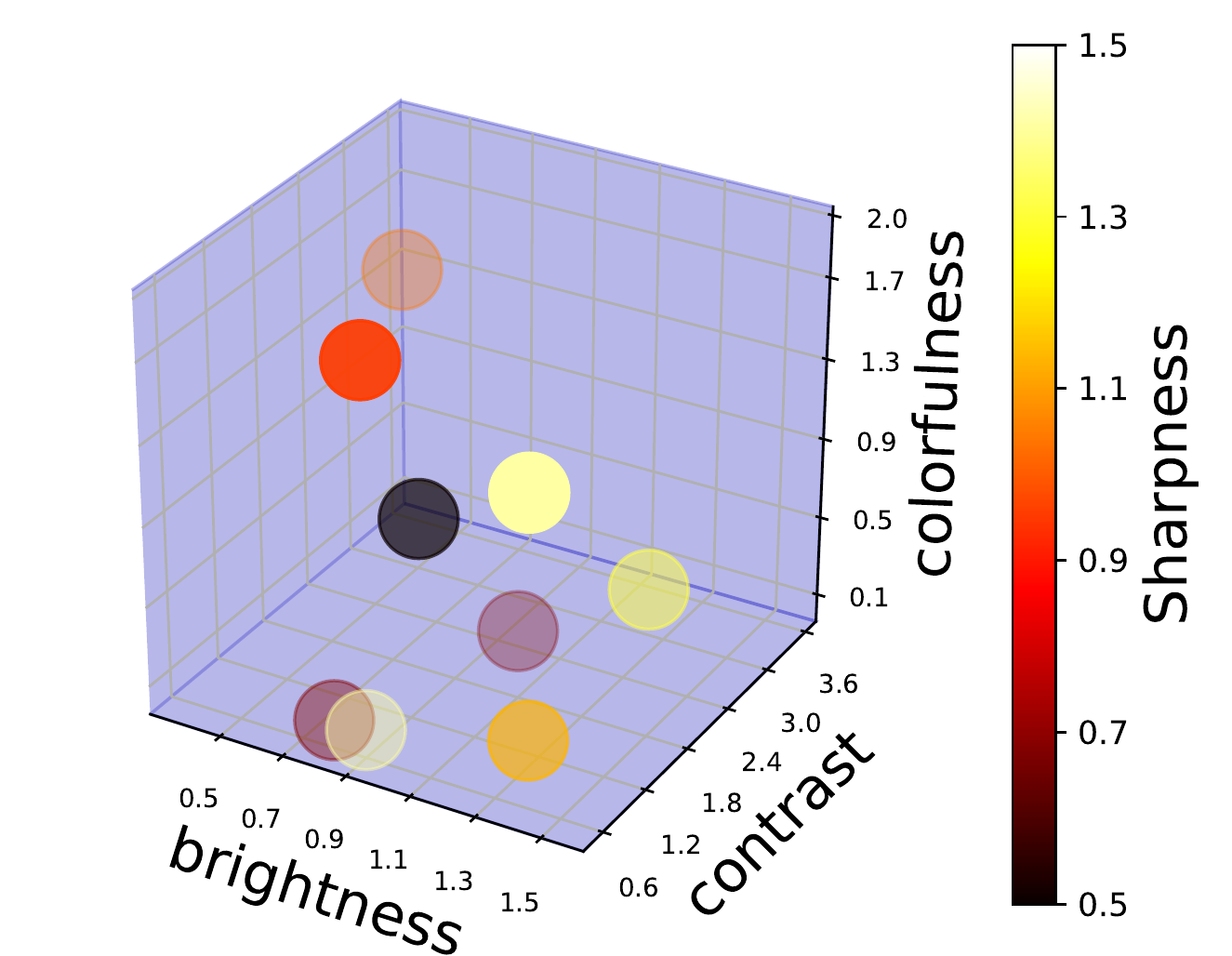}
    \caption{Face detection}
    \label{fig:olympics_face}
\end{subfigure}%
\begin{subfigure}[]{0.5\linewidth}
\centering
    \includegraphics[height=1.3 in]{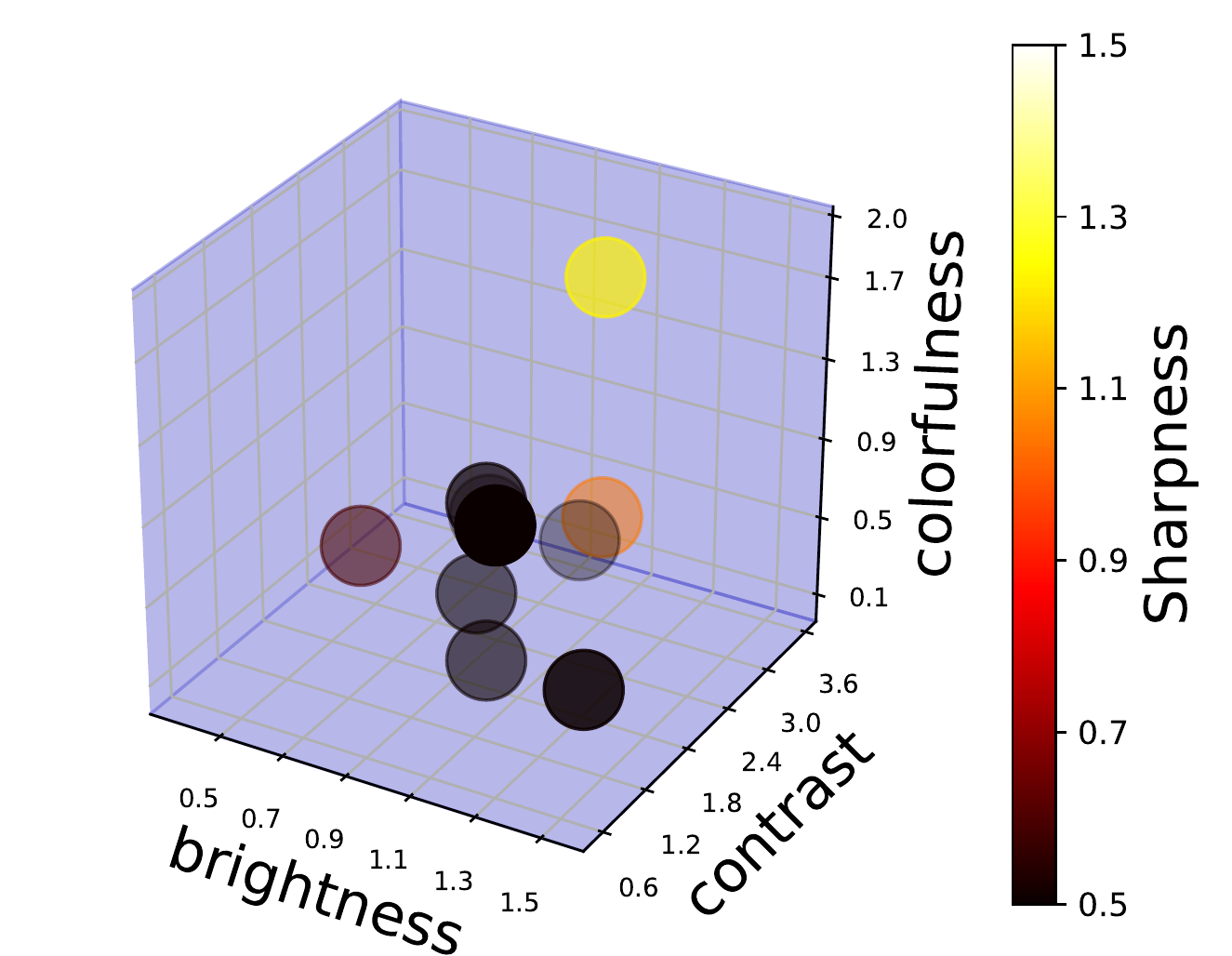}
    \caption{Person detection}
    \label{fig:olympics_person}
\end{subfigure}
\caption{Distribution of best \emph{transformation tuple} for two AUs on \emph{Olympics} video snippets.}
\label{fig:olympics}
 \vspace{-0.2in}
\end{figure}

We consider 19 video snippets from the HMDB dataset~\cite{Kuehne11}
and 11 video snippets from the Olympics
dataset~\cite{niebles2010modeling} that capture different content under different environmental conditions while using the default camera parameter. Using cvat-tool~\cite{cvat}, we
manually annotated the face and person bounding boxes to form our
ground truth.  Each video-snippet contains no more than a few hundred
frames, 
and the environmental conditions 
vary across the video snippets {due to change in
  video capture locations}. We determine a single best tuple of
those four transformations for each video, \ie one that results in the highest
analytical quality for that video.
\figref{fig:hmdb51} and \figref{fig:olympics} show the distribution of
the best transformation tuples for the videos in the two datasets,
respectively. We see that with a few exceptions, the best
transformation tuples for different videos (\ie that capture different content under various environmental condition) in a dataset do not
cluster, suggesting that any fixed real camera parameter settings will
not be ideal for different environmental conditions or input content as well as it also varies for different analytics
tasks.  \tabref{tab:best-video-conf} shows the maximum and average
analytical quality improvement achieved after transforming each
video-snippet as per their best transformation tuple. We observe 
up to 58\% improvement in accuracy of insights when
appropriate transformations
  or equivalent camera parameters 
are applied.

%\comment{where is Table 1 cited???}

\begin{table}[tb]
\begin{center}
\vspace{-0.15in}
\caption{Accuracy improvement of best configuration.}
\label{tab:best-video-conf}
{
\small
\begin{tabular}{||c|c||c|c|c|}
    \hline
    Video-Dataset & AU & \multicolumn{2}{c|}{mAP}\\
    & & \multicolumn{2}{c|}{improvement}\\
    \cline{3-4}
    & & Max & Mean \\
    \hline
    \multirow{2}{*}{Olympics} & Person Detection & 40.38 & 8.38 \\
    \cline{2-4}
    & Face Detection  & 19.23 & 1.68  \\
    \hline
    \multirow{2}{*}{HMDB} & Person Detection & 57.59 & 12.63 \\
    \cline{2-4}
    & Face Detection  & 18.75 & 4.22  \\
    \hline
\end{tabular}
}
\end{center}
\vspace{-0.1in}
\end{table}

%\cut{
In summary, 
environmental changes and input content variations can result in low-quality image acquisition, which in
turn result in poor analytics accuracy. Tuning the camera parameter
settings during capture can provide improvement in accuracy of
AUs, but such camera parameter
tuning is hard for a human to do manually because the best
parameter combination will vary with location of the camera, the type of
analytics units, and the environmental conditions.
%  Humans cannot keep
% adjusting these parameters manually, 
This calls for developing methods 
that can automatically adapt the camera parameters
to improve the accuracy of AUs.
%}

% We discuss the challenges and our approach in designing such a system
% design in the next section.

\vspace{-0.05in}
\section{Challenges}
\label{sec:challenge}

In this paper, we propose to develop a camera tuning framework 
that dynamically adapts the four parameter settings of the video-capturing camera
in a video analytics system (VAS) to optimizes the accuracy of its AUs.
Designing such a framework faces two challenges. 
Below, we discuss these two challenges and our approaches to addressing each one of them.

{\bf Challenge 1: Identifying the best camera settings for a
  particular scene.} Identifying the best camera settings for a given
scene that gives the best AU accuracy is challenging even during
offline, as it requires comparing the impact of all possible camera
settings. Doing so in an online manner is even more challenging.

\textbf{Approach.} To address this challenge, we propose to
use an online learning method. Particularly, we use Reinforcement
Learning (RL)~\cite{introduction-to-rl}, in which the agent learns
the best camera settings on-the-go. 
Using RL, we do not have to know
apriori the various scenes that the camera would observe. Instead, the
RL agent learns and identifies automatically the best camera
settings that give the highest AU accuracy for any particular
scene.
Out of several recent RL algorithms,
we choose the SARSA~\cite{q-lambda} RL algorithm for identifying the best
camera settings.

While RL is a fairly standard technique, applying it to tuning camera parameters in a real-time video analytics system in turn raises one unique challenge as follows.

{\bf Challenge 2: No Ground truth in real time.} Implementing the
online RL approach requires knowing the quality (\ie either reward or
penalty) of every action taken during exploration and
exploitation. Measuring the quality of camera parameters' change in
absence of ground-truth is challenging.

\textbf{Approach.} We proposed to leverage state-of-the-art perceptual
Image Quality Assessment (IQA) methods as a proxy of the quality
measure (\ie the reward function). Specifically, we experimentally evaluate a list of 
state-of-the-art IQA methods and use the best-performing one as the reward function
in the RL engine.
\if 0
RankIQA~\cite{liu2017rankiqa} perceptual quality estimator as the
quality estimator of actions taken by RL agent. \comment{Why do we use
  this one? Is this the only chocie?} \response{The impact of
  different SOTA perceptual IQA methods on camera parameter tuning is
  further discussed in~\secref{subsec:diff-IQA}}
\fi

\vspace{-0.05in}
\section{Design}
\label{sec:design}

\begin{figure}[tp]
   \centering \includegraphics[width=1.02\linewidth]{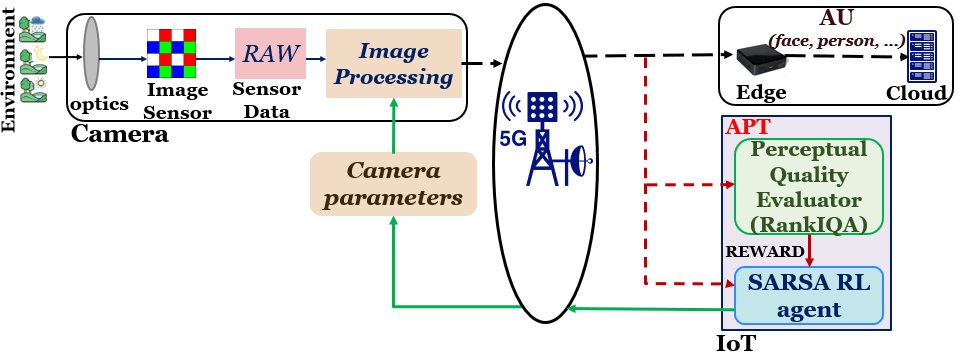} 
   \caption{$\approach$ system design.} 
   \vspace{-0.2in}
\label{fig:system_design}
   %\vspace{-0.1in}
\end{figure}

~\figref{fig:system_design} shows the system-level architecture for
\approach, which automatically and adaptively tunes the camera
parameters to optimize the analytics accuracy. \approach\ incorporates
two key components: a
perceptual no-reference quality estimator and a Reinforcement Learning (RL) engine. 

%  For \approach, we use
%  \emph{rankIQA}~\cite{liu2017rankiqa} as our perceptual quality
%  estimator. More discussion on specific IQA selections are presented
%  in~\secref{subsec:diff-IQA}.

%explain the architecture in detail also

\subsection{Perceptual No-reference Quality Estimator}
\label{perceptual-iqa}
Since it is not possible to obtain ground-truth in real-time to measure the accuracy of video analytics applications, we rely on a technique which won't require ground truth, but still help in improving the analytics accuracy. To this end, we leverage SOTA perceptual no-reference quality estimator, which gives an estimate of the quality of the frame produced by the video camera. Better the quality of the frame, better will the analytics be able to generate accurate insights. Therefore, in the design on \approach\ we employ such IQA method to help guide the system in choosing appropriate camera settings. We discuss our choice of IQA method in \secref{subsec:diff-IQA} and how this IQA method is used in RL engine within \approach\ in \secref{camera-parameters-tuning}.

%In absence of ground-truth in real-time, RL engine needs to know whether its actions are impacting the capture quality in the positive or negative direction. We use SOTA no-reference (blind) IQA method as a guide and generates reward/penalty for the RL agent. More discussion on several SOTA no-reference IQA method can be found in~\secref{subsec:diff-IQA}.

\subsection{Reinforcement Learning (RL) Engine}
\label{camera-parameters-tuning}

RL engine is the heart of \approach, as it is the
one that automatically chooses the best camera settings for a
particular scene.  In designing the RL engine, we considered popular
RL algorithms such as Q-learning~\cite{q-learning} and
SARSA~\cite{q-lambda} which are general techniques and highly
effective in learning the best action to take in order to maximize the
reward.  To choose between the two options, we experimentally compared
them in the context of choosing the best camera settings and found that
training with SARSA achieves slightly faster convergence than with
Q-learning. Therefore, we decide to use the SARSA RL algorithm in \approach.

%Comparing the state space and the way actions are chosen in these two algorithms, Q-learning can potentially take certain ``dangerous actions along the cliff'' {which can lead to large negative rewards for the AU.} SARSA, on the other hand, avoids such dangerous actions. Therefore, although in our case, we do not see such cliffs, where sudden large negative rewards are obtained for certain actions, we choose to be more conservative and go with SARSA algorithm. 

Like other RL algorithms, in SARSA, an agent continuously interacts
with the environment (\textit{state}) it is operating in, by taking
different \textit{actions}. As the agent takes an action, it moves into
a new state or environment. For each action, there is an associated
\textit{reward} or penalty, depending on whether the new state is
more desirable or not. Over time, as the agent continues taking
actions and receiving rewards and penalties, it learns to maximize the
rewards by taking the right actions, which ultimately lead the agent
towards desirable states.

%To automatically tune the camera parameters, we use reinforcement learning (RL) \cite{introduction-to-rl} techniques. We use State-Action-Reward-State-Action (SARSA) RL algorithm~\cite{q-lambda}. In this setup, there is an agent, which interacts with the environment (\textit{state}) it is in, by taking different \textit{actions}. As the agent takes actions, it moves into a new state or environment. For each action, there is an associated \textit{reward} or penalty, depending on whether the new state is desirable or not Over a period of time, as the agent continues taking actions and receiving rewards and penalties, the agent learns to maximize the rewards by taking the right actions, which ultimately leads the agent towards desirable states.

%SARSA does not require any labeled data or pre-trained model, rather it learns and updates its policy based on the action it takes. Therefore, it is an on-policy learning algorithm and requires a clear definition of the environment (state), actions and reward. For the purpose of automatic camera-parameter tuning, we define them as follows:

As with many other RL algorithms, SARSA does not require any labeled
data or pre-trained model, but it does require a clear definition of
the \textit{state}, \textit{action} and \textit{reward} for the RL
agent. This combination of \textit{state}, \textit{action} and
\textit{reward} is unique for each application and therefore needs to
be carefully chosen, which ensures that the agent learns exactly what is
desired. In our setup, we define them as follows:

%\textit{\underline{State}}: A state is a tuple of two vectors, $s=<P_{j}, V_{s_{j}}>$, where $P_j$ consists of the current brightness, contrast, sharpness and color parameter values on the camera, and $V_{s_j}$ consists of measure of brightness, contrast, sharpness and color-saturation of the captured frame.

\textit{\underline{State}}: A state is a tuple of two vectors,
$s_t=<P_{t}, M_{t}>$, where $P_t$ consists of the current brightness,
contrast, sharpness, and color-saturation parameter values on the
camera, and $M_{t}$ consists of the measured values of brightness,
contrast, color-saturation, and sharpness of the captured frame at
time $t$, measured as
in~\cite{bezryadin2007brightness,peli1990contrast,hasler2003colormeasuring,
  de2013sharpness}.

\textit{\underline{Action}}: The set of actions that the agent can
take are (1) to increase or decrease one of the brightness, contrast,
sharpness or color-saturation parameter value, or (2)
not to change any parameter values.

\textit{\underline{Reward}}: We use the best-performing (experimentally chosen as described in ~\secref{subsec:diff-IQA})
% \emph{rankIQA} perceptual quality estimator~\cite{liu2017rankiqa} 
perceptual quality estimator
as the immediate reward function (r) for the SARSA
algorithm.
Along with considering immediate reward, the agent
  also factors in future reward that may accrue as a result of the
  current actions. Based on this, a value, termed as Q-value (also
  denoted as $Q(s_t,a_t)$) is calculated for taking an action $a_t$ when in
  state $s_t$ using Equation~\ref{eq:sarsa}.
\begin{equation}
%\vspace{-0.05in}
\label{eq:sarsa}
Q(s_t,a_t) \leftarrow Q(s_t,a_t) +
\alpha \left[ r + \gamma \cdot Q(s_{t+1}, a_{t+1}) - Q(s_t,a_t) \right]
\end{equation}
Here, $\alpha$ is learning rate (a constant between 0 and 1) used to control how
much importance is to be given to new information obtained by the
agent. A value of 1 will give high importance to the new information
while a value of 0 will stop the learning phase for the agent.
%During the initial phase, a higher value e.g. 0.85 can be assigned by the operator, which will help the agent quickly \textit{learn} and understand the new environment by giving higher importance to the most recent information. After the agent has learnt and assimilated the information from the new environment, the value of $\alpha$ can be lowered e.g. 0.1 by the operator so that the agent now starts \textit{using} the gathered information about the new environment.

Similar to $\alpha$, $\gamma$ (also known as
the discount factor) is another constant used to control the importance given by the agent
to any long term rewards. A value of 1 will give very high importance
to long term rewards while a value of 0 will make the agent ignore any
long term rewards and focus only on the immediate rewards. 
If the conditions do not change frequently, a higher value, \eg
0.9, can be assigned to prioritize long term rewards;
if the environmental conditions change very frequently, a lower value,
\eg 0.1, can be assigned to $\gamma$ to prioritize immediate rewards.

{\bf Exploration vs. Exploitation.}
We define a constant called $\epsilon$ (between 0 and 1) 
to control the balance between exploration vs. exploitation when the agent 
takes actions.
In particular, at each step, the agent generates a random number
between 0 and 1; if the random number is greater than the set value of
$\epsilon$, then a random action (exploration) is chosen, else it performs 
exploitation.

\vspace{-0.05in}
\section{Evaluation}
\label{sec:eval}

We first evaluate several design choices for the perceptual IQA method to be used
as the reward function in the RL engine in terms of 
their impact on analytics performance in a controlled mock-up scene
(\secref{subsec:diff-IQA}), and pick the best-performing choice for use in \approach.
We then extensively evaluate the
effectiveness of \approach\ on the mock-up scene under different initial
parameter settings (\secref{subsec:react}) and in a real-world
%parking-lot VAS 
deployment (\secref{subsec:real-world}).

\subsection{Effect of using different IQA Methods}
\label{subsec:diff-IQA}

Throughout the last decade, several no-reference (blind) IQA
methods~\cite{brisque_mittal2012no,biqi_moorthy2010two,gmlog_xue2014blind,kang2014convolutional,Su_2020_CVPR,liu2017rankiqa}
have been proposed to improve the video/image capture quality based on human
perception. In this section, we evaluate the impact of three different
blind IQA methods that are designed to estimate the quality of real-world
distorted images for use as \approach's quality evaluator. Since
downstream analytics focus on low-level local features (\ie extracted
via convolution layers) for deriving insights from the input video
stream, we choose three popular perceptual IQA methods that employ
convolution network.
%as feature extraction.

CNN-IQA~\cite{kang2014convolutional} is the first to use the spatial domain
without relying on hand-craft features used by previous IQA
methods. It combines feature learning and quality regression
in one optimization process which leads to a more effective quality
estimation model.  Hyper-IQA~\cite{Su_2020_CVPR} decouples the IQA
procedure into three stages: content understanding, perception rule
learning and finally quality prediction. {Hyper-IQA} estimates image
quality in a self-adaptive manner by adaptively running 
different hyper-networks.  Finally,
Rank-IQA~\cite{liu2017rankiqa} addresses the problem of limited size
of the IQA dataset during training. It uses a siamese network to rank
images and then uses the ranked images to train deeper and wider
convolution networks for absolute quality prediction.

To assess the impact of using different IQA estimators as reward functions on the analytics
performance under the same environmental conditions, we use a mock-up scene
with a fixed number of objects (\ie cars and persons). In this mock-up
scene, 3D slot cars are continuously moving along the track and 3D
human models are kept stationary. In doing so, this experimentally controlled
mock-up scene provides controllability and replicability in
experimental setup and enables us to try out different reward functions under the same
environment and content.

\begin{figure} [tb]
\if 0
\begin{subfigure}[t]{0.3\linewidth}
\centering
    \includegraphics[width=\linewidth]{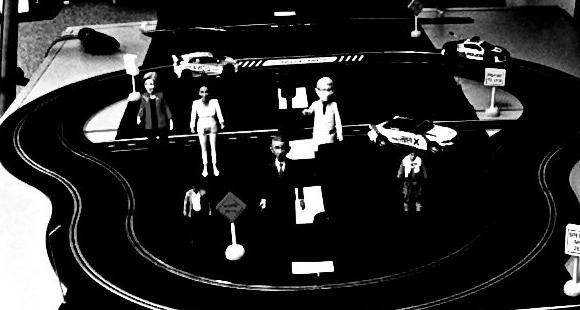}
    \caption{capture under S1}
    \label{fig:static-stream-1}
\end{subfigure}
\fi
\begin{subfigure}[t]{0.48\linewidth}
\centering
    \includegraphics[width=\linewidth]{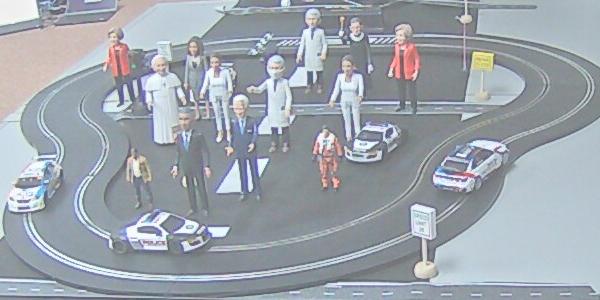}
    \caption{capture under fixed setting}
    \label{fig:static-stream-2}
\end{subfigure}
\begin{subfigure}[t]{0.48\linewidth}
\centering
    \includegraphics[width=\linewidth]{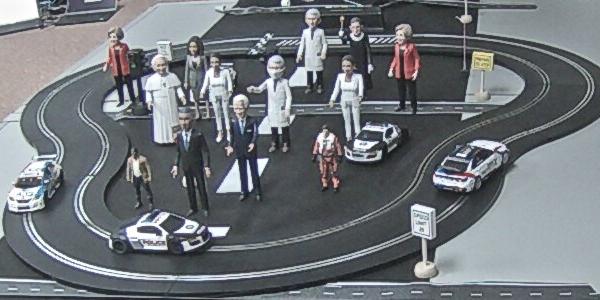}
    \caption{\approach camera capture}
    \label{fig:camtuner-stream}
\end{subfigure}%
   % \vspace{-0.1in}
 \caption{Sample camera captures.}
  \label{fig:camtuner-static-stream}
 \vspace{-0.1in}
\end{figure}

We first train \approach\ using each of these three quality estimator
output as the reward function for one hour on the mock-up 
scene.
%directly changing network camera parameters. 
During training, after every 2 minute interval, we 
change the camera parameters to emulate different
environmental conditions.
For evaluation, we placed two identical \emph{AXIS 3505 MK-II network cameras} side-by-side in front of the
mock-up scene as shown in~\figref{fig:camtuner-static-stream}.  During
evaluation, we used 5 different camera settings and observed how
\approach\ reacts to those initial camera parameter settings.

\begin{figure} [tb]
    \centering
    \begin{subfigure}[t]{\columnwidth}
        \vskip 0pt
        \centering
        \includegraphics[width=0.49\textwidth]{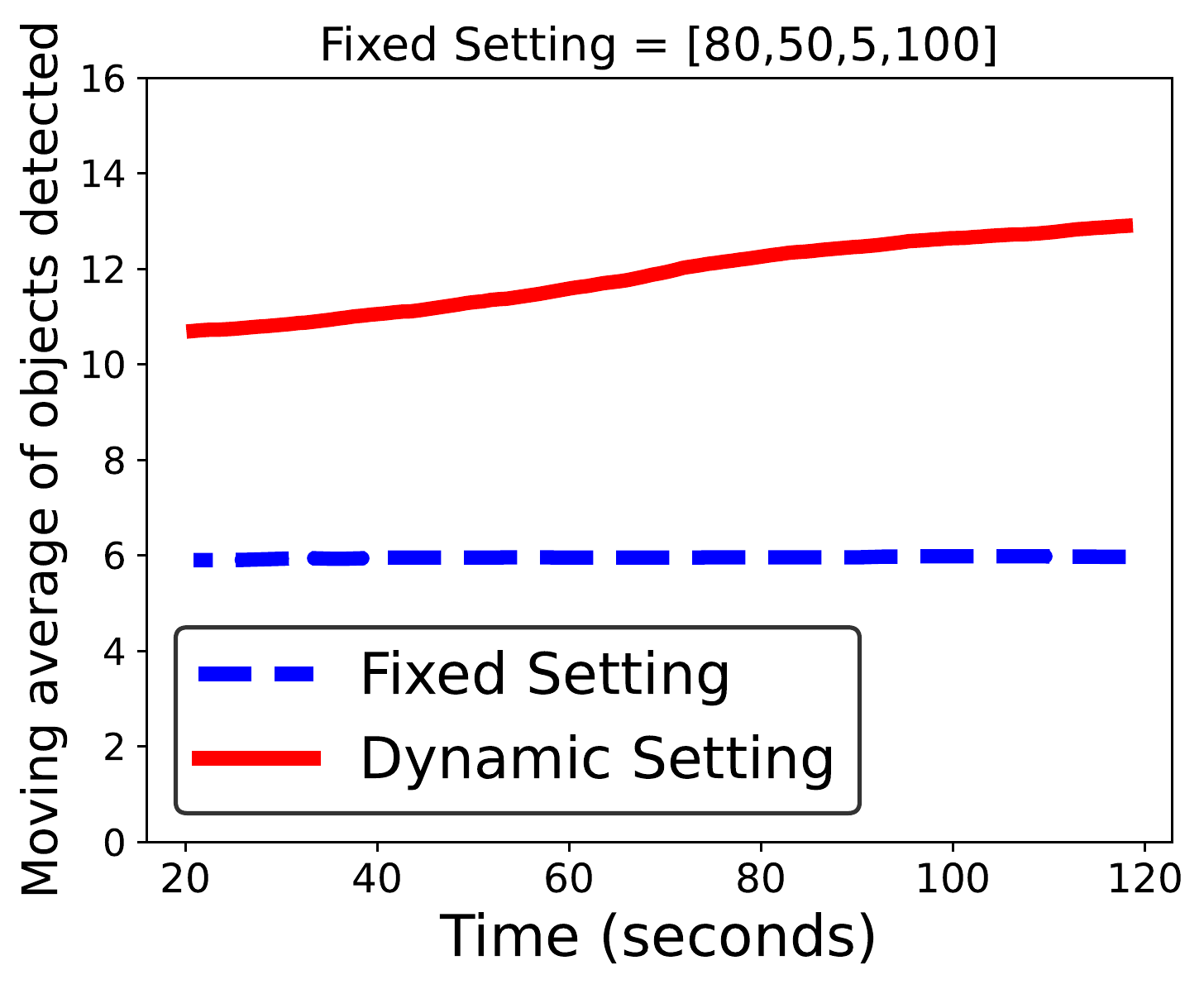}
        \includegraphics[width=0.49\textwidth]{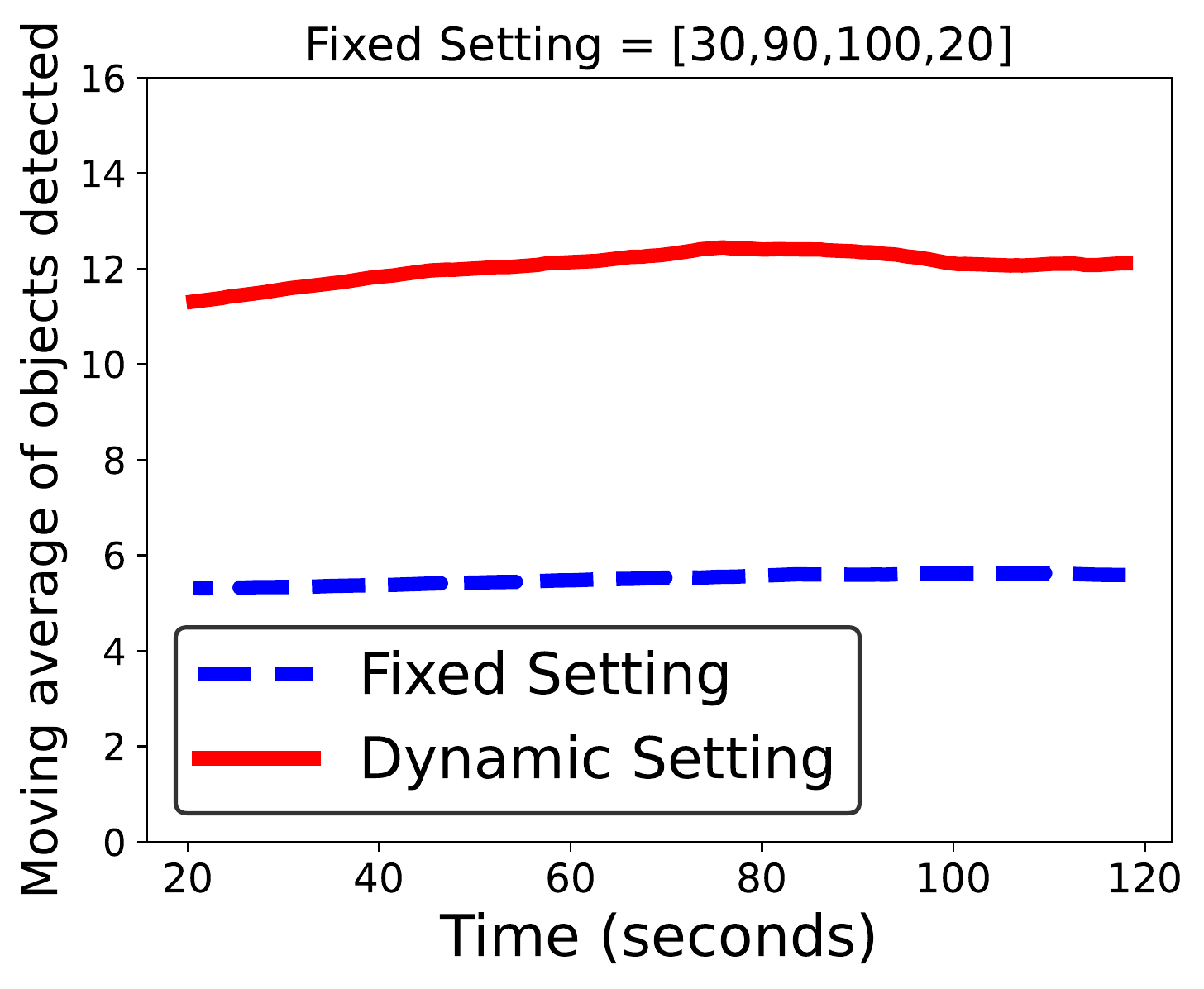}
        \caption{HyperIQA as reward}
        \label{fig:hyperIQA}
    \end{subfigure}
        \hfill
        \begin{subfigure}[t]{\columnwidth}
        \vskip 0pt
        \centering
        \includegraphics[width=0.49\textwidth]{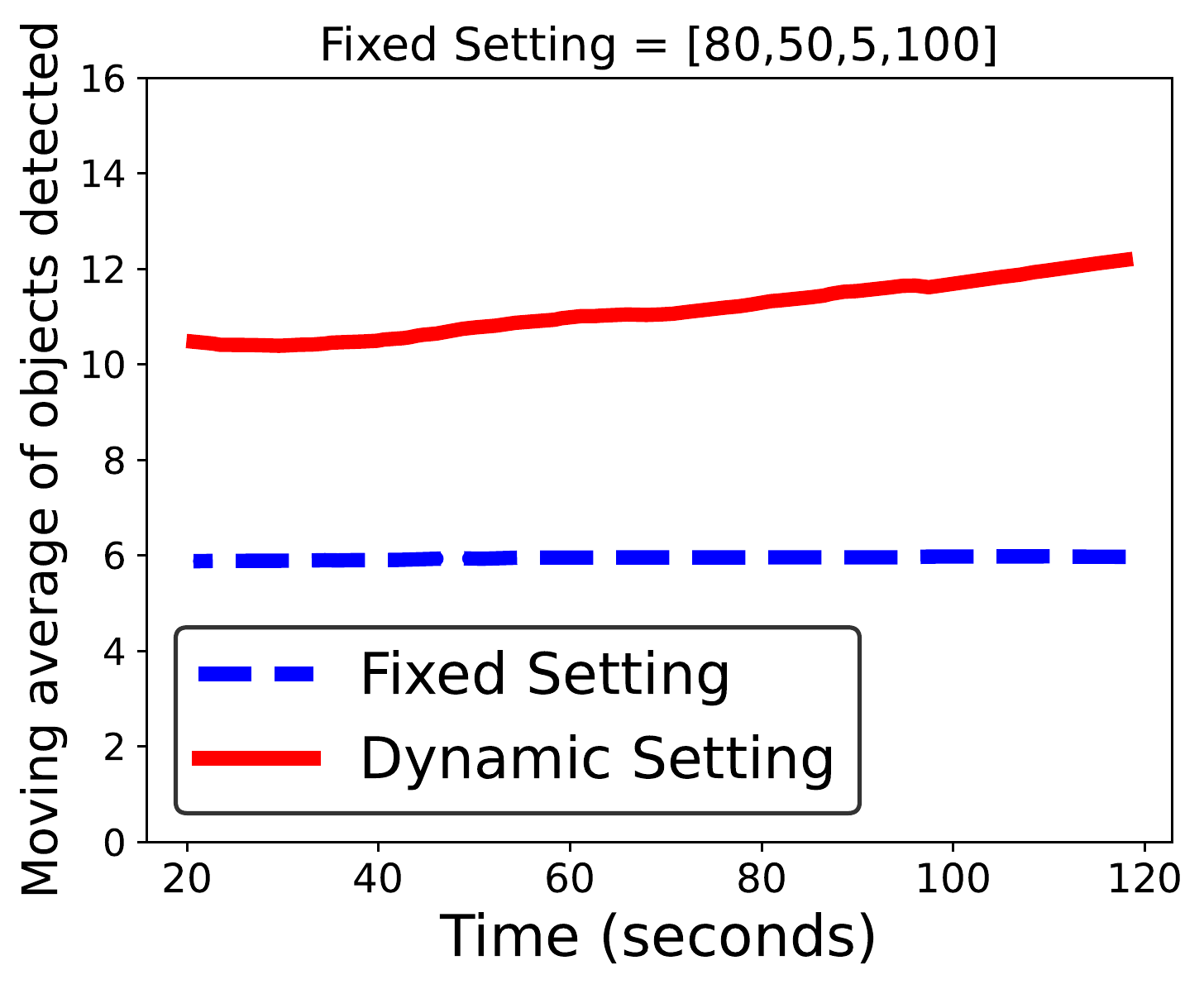}
        \includegraphics[width=0.49\textwidth]{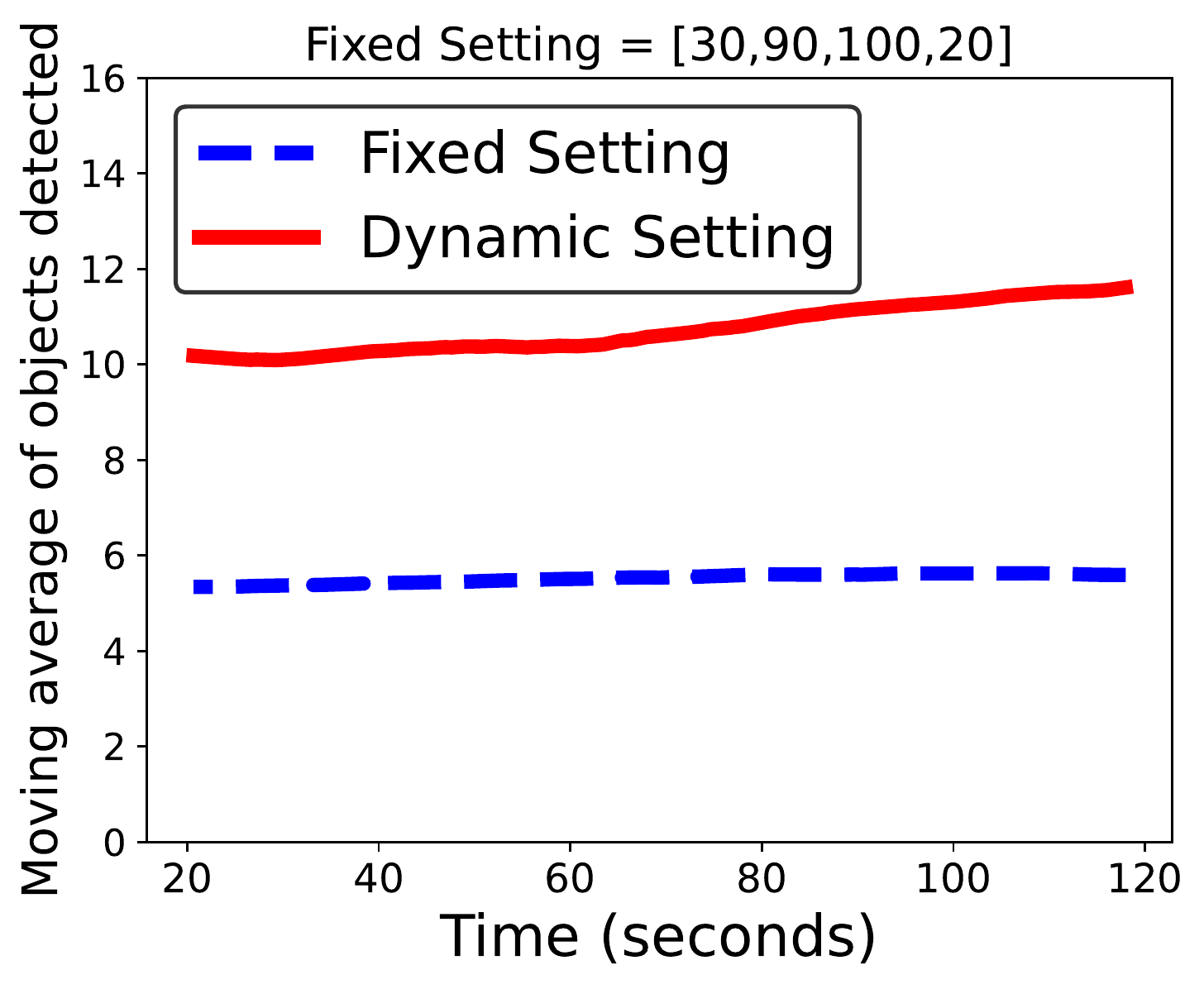}
        \caption{CNNIQA as reward}
        \label{fig:cnnIQA}
    \end{subfigure}
        \hfill
    \begin{subfigure}[t]{\columnwidth}
        \vskip 0pt
        \centering
        \includegraphics[width=0.49\textwidth]{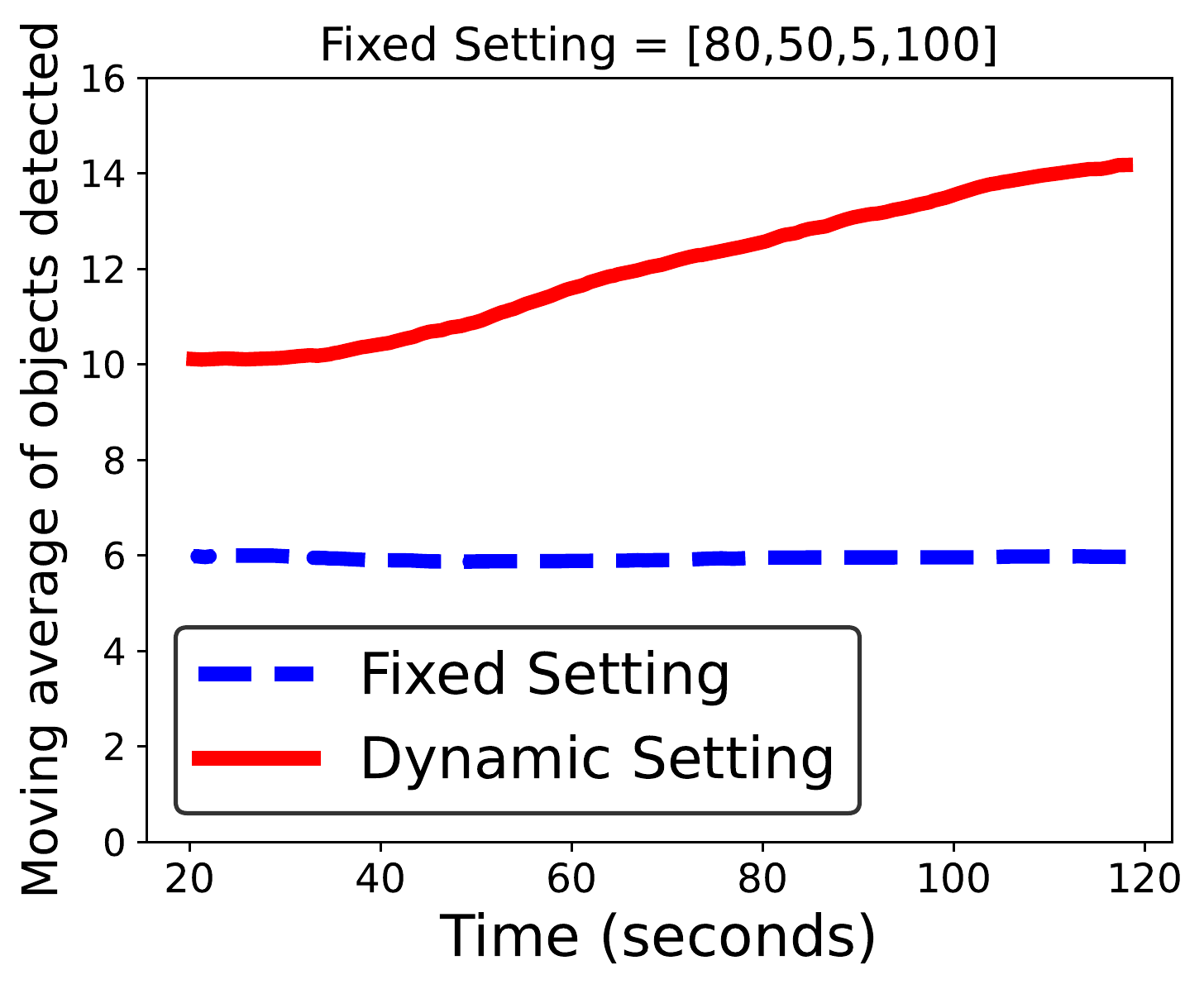}
        \includegraphics[width=0.49\textwidth]{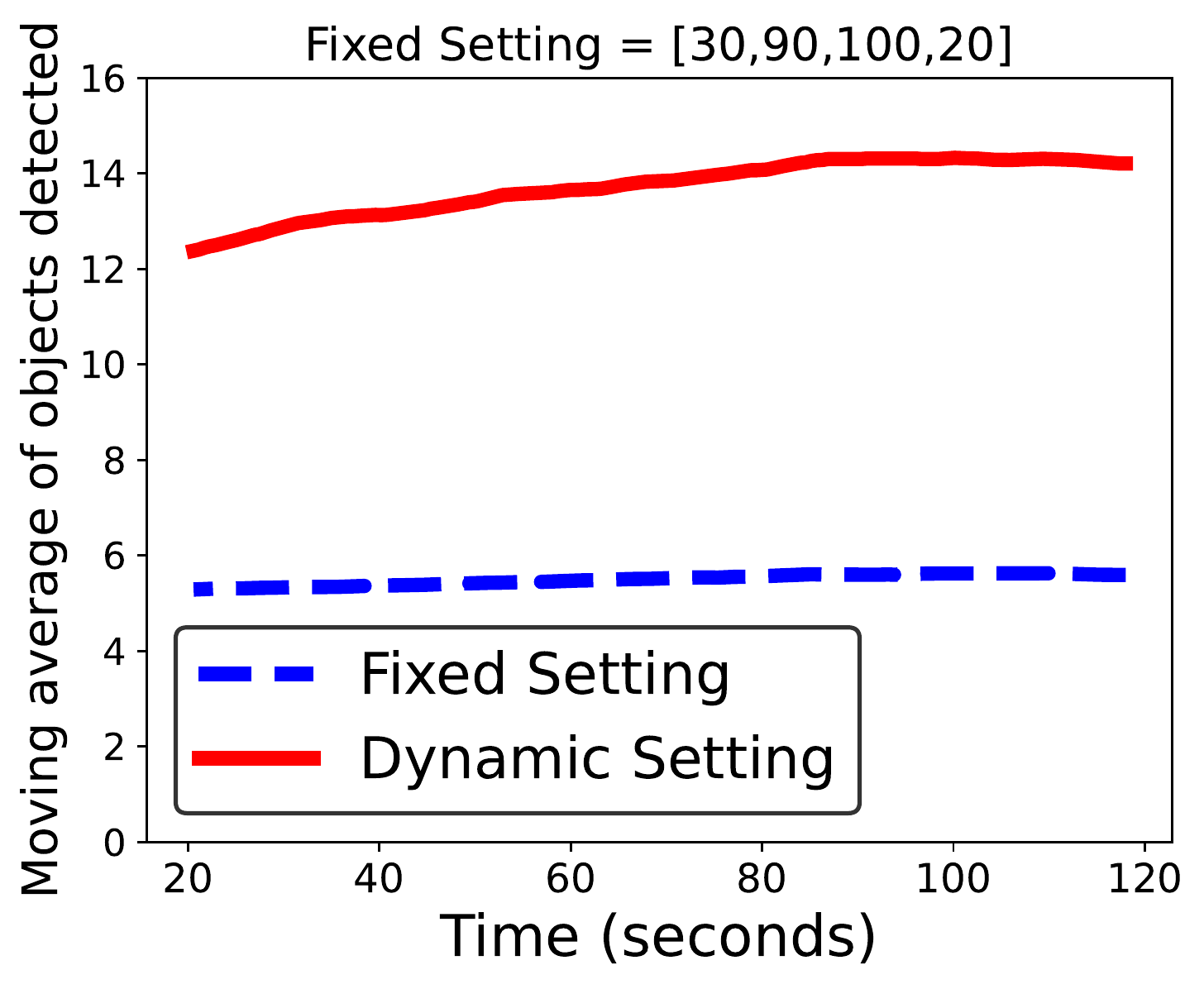}
        \caption{RankIQA as reward}
        \label{fig:rankIQA}
    \end{subfigure}
\caption{\approach reaction to different initial camera settings under different IQA metrics as reward function (Moving average of per-frame object detection is computed over a window of last 100 frames, shown in Y axis.)}
\vspace{-0.1in}
\label{fig:eval_diff_IQA_suboptimal}
\end{figure}

~\figref{fig:eval_diff_IQA_suboptimal} shows how the three different IQA
methods effectively guide SARSA RL agent in \approach, resulting in higher true-positive object detections 
%than the 
when compared to object detector's performance on the stream with fixed camera
setting.
~\tabref{tab:IQA-impact-table} presents the average
improvement in true-positive object detections observed throughout
multiple 2-minute time-intervals, and the average number of objects
detected in the steady state for the three different reward functions.
We observe that \emph{Rank-IQA} guides SARSA-RL agent better under
environmental variations which in turn leads to more object
detections from the same scene.
%  and higher improvement in object detections. 
Thus, we use Rank-IQA as our perceptual quality estimator
for \approach.

%\subsection{How quickly does \approach react to different camera settings}
\subsection{Effectiveness of \approach\ in a Mock-up Scene}
\label{subsec:react}

\begin{table}[tb]
\centering
\caption{Comparing IQA methods as reward functions}
\begin{tabular}{||c|c|c||}
\hline
IQA & Improv. over fixed settings& Objs detected     \\ 
& (Avg) \% & steady state (Avg) \\ \hline\hline
\textbf{Hyper-IQA} & 132.6 & 13.5\\ \hline
\textbf{CNN-IQA} & 141.2 & 14.2 \\ \hline
\textbf{Rank-IQA} &  150.5 & 15.1  \\ \hline
\end{tabular}
\vspace{-0.2in}
\label{tab:IQA-impact-table}
\end{table}

\begin{figure}
    \centering
    \begin{subfigure}[t]{0.48\columnwidth}
        \vskip 0pt
        \centering
        \includegraphics[width=0.99\textwidth]{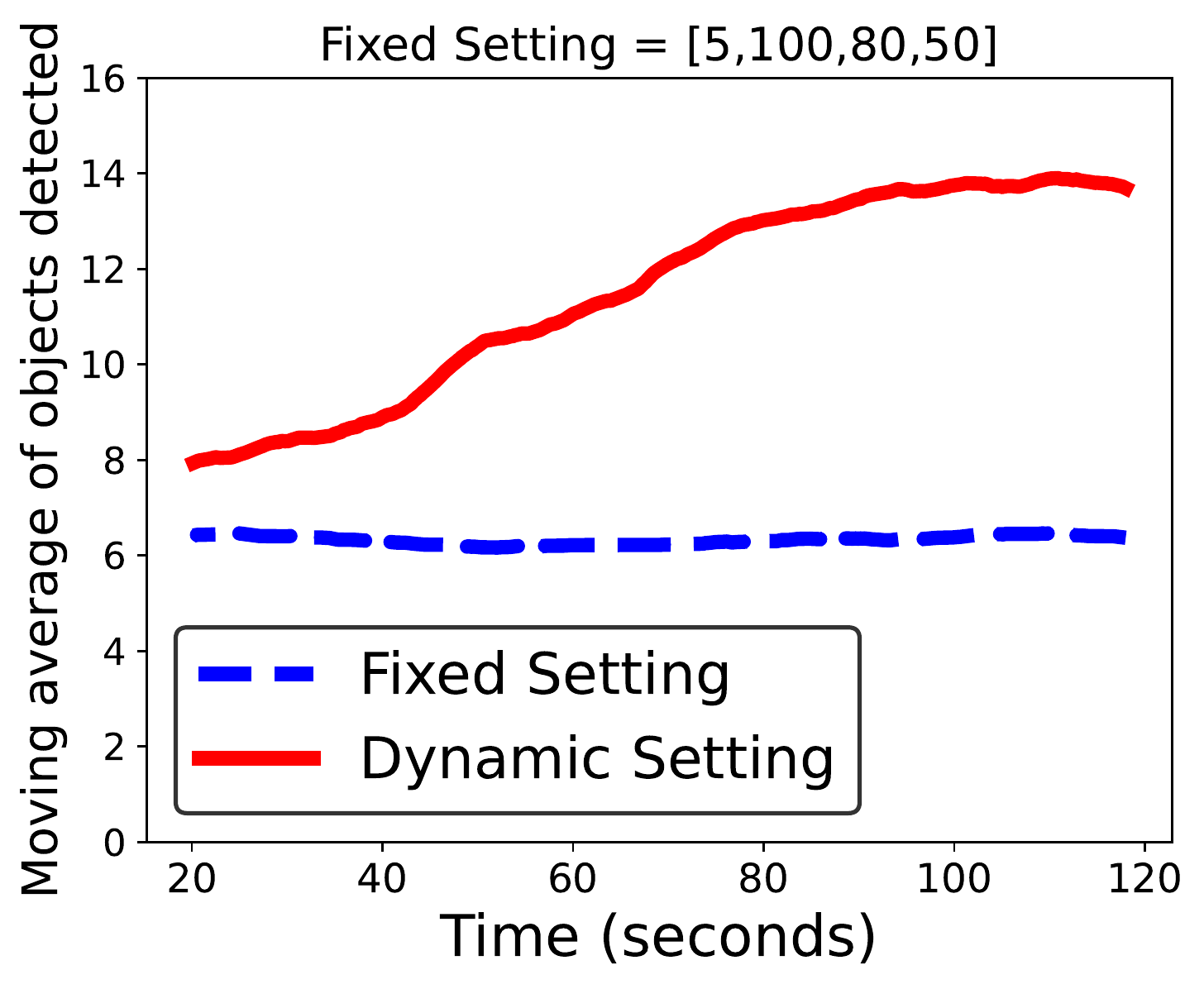}
        \caption{Fixed setting 1 (S1)}
        \label{fig:suboptimal_1}
    \end{subfigure}
    \hfill
    \begin{subfigure}[t]{0.48\columnwidth}
        \vskip 0pt
        \centering
        \includegraphics[width=0.99\textwidth]{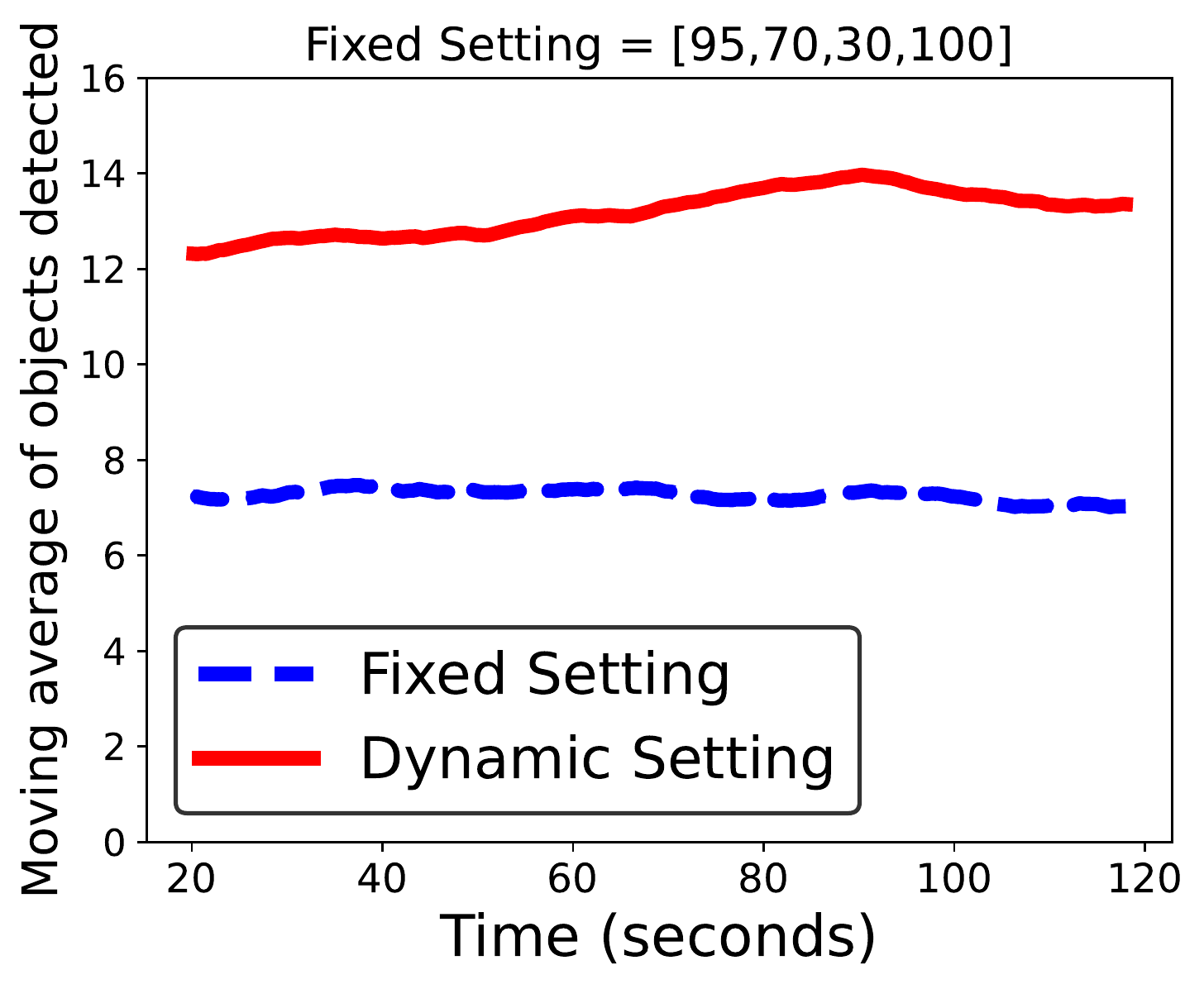}
        \caption{Fixed setting 2 (S2)}
        \label{fig:suboptimal_2}
    \end{subfigure}
    \hfill
    \begin{subfigure}[t]{0.48\columnwidth}
        \vskip 0pt
        \centering
        \includegraphics[width=0.99\textwidth]{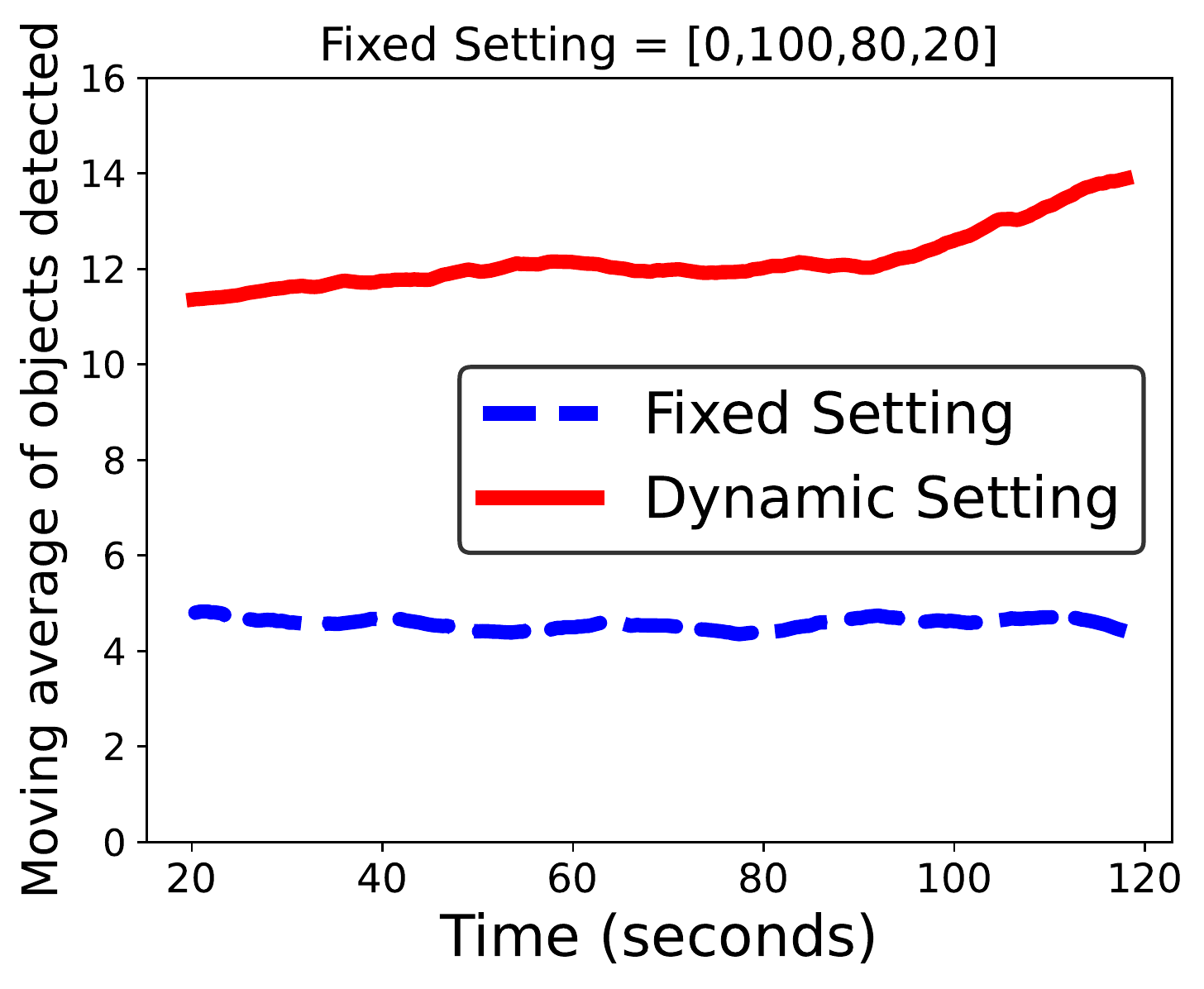}
        \caption{Fixed setting 3 (S3)}
        \label{fig:suboptimal_3}
    \end{subfigure}
     \hfill
        \begin{subfigure}[t]{0.48\columnwidth}
        \vskip 0pt
        \centering
        \includegraphics[width=0.99\textwidth]{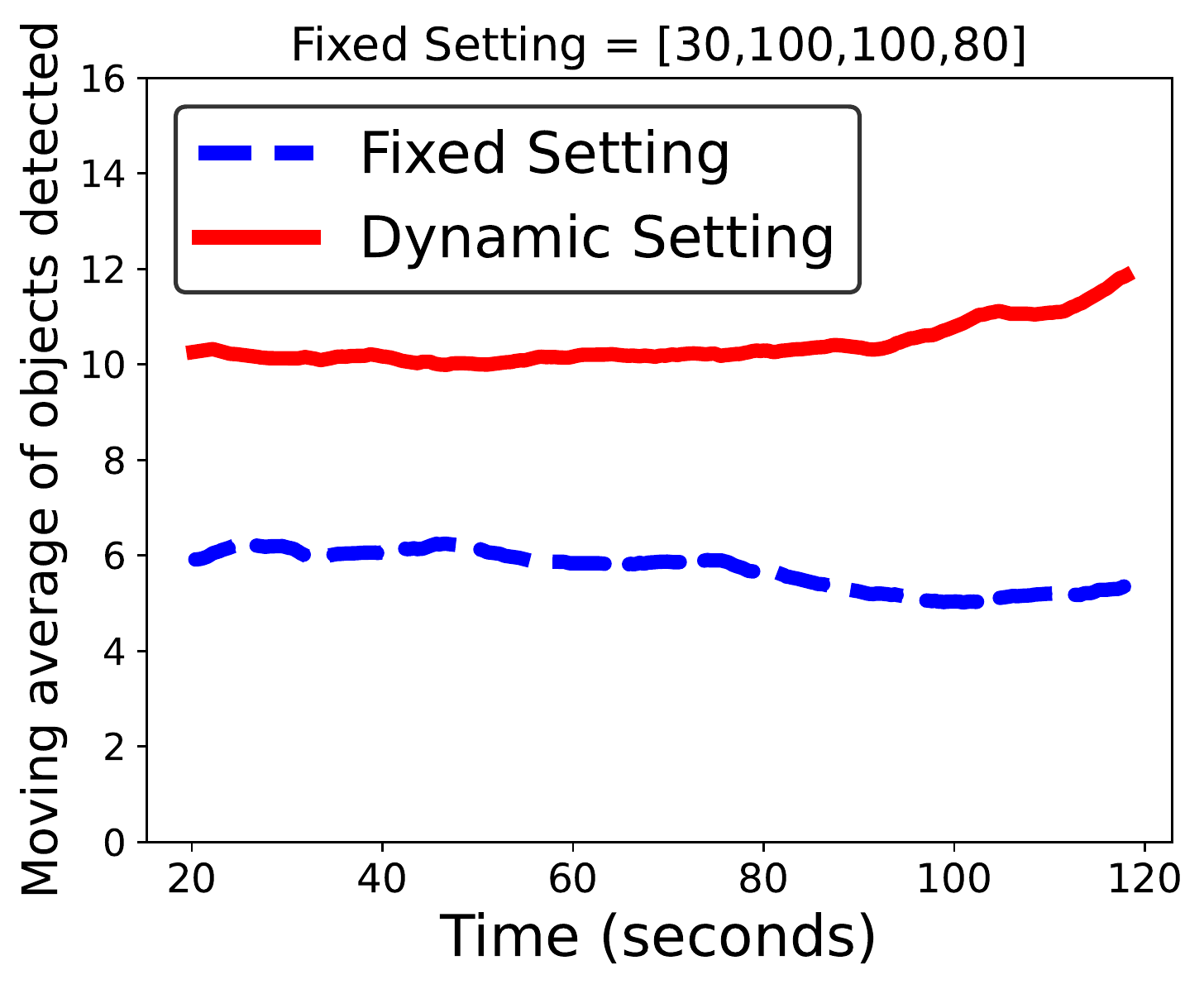}
        \caption{Fixed setting 4 (S4)}
        \label{fig:suboptimal_4}
    \end{subfigure}
\caption{\approach reaction to different camera settings using Rank-IQA reward (moving average of per-frame object detection is computed over a window of past 100 frames, shown in the Y-axis.)}
\label{fig:eval_suboptimal}
\vspace{-0.2in}
\end{figure}

Here, we evaluate how quickly \approach can react to any initial
setting and converge to a setting that can provide better
analytical outcome. We use the same controlled mock-up scene described
in \secref{subsec:diff-IQA}.
Both cameras start with same initial setting (we use four different
camera settings denoted as S1, S2, S3 and S4, respectively) and stream at 10 FPS
over a 2-minute period, during which the four parameters
of Camera 1 are kept to the same initial values, while the parameters of Camera 2 are tuned dynamically by \approach every 2 seconds.
{On every frame streamed from the camera}, we use
Yolov5~\cite{glenn_jocher_2022_6222936} object detector to
detect objects and record the type of objects with their bounding
boxes~\footnote{Manual inspection confirms there is no false-positive detection
  in the 2-minute period.}.
~\figref{fig:eval_suboptimal} shows the moving average of per-frame
object detections (with a window size of 100 frames) in the Y axis (to clearly
show the trend) of the two cameras under four different initial
%suboptimal 
settings.
We observe there is an initial gap between the performance of YOLOv5
between the two camera streams which indicates that within the first
10 seconds, \approach changes the camera parameters based on
human-perceptual quality estimator (\ie \quality) output and achieves
better object detection.  Furthermore, we observe that \approach
gradually finds best-possible setting within one minute that
enables Yolov5 to detect more number of objects from the captured scene
(total 4-9 more object detections per frame compared to detections on camera stream with fixed setting).

%\subsection{Real-world Deployment (Parking lot)}
\subsection{Effectiveness of \approach\ in Real-world Deployment}
\label{subsec:real-world}
To evaluate the effectiveness of camera parameter tuning by
\approach\ in a real-world deployment, we use two co-located
\emph{AXIS Q3515} network cameras that continuously monitor an
enterprise parking lot.
%as shown in~\figref{fig:real_setup}
%While one camera is set to manufacturer-provided default parameters, \approach\ tunes the four image-appearance parameters of the other camera. 
Here, one camera is set to manufacturer-provided default settings, while the other camera parameters are adaptively tuned by \approach. Captured frames from each camera are uploaded to a remote
edge-server (equipped with Intel-Xeon processor and NVIDIA Geforce
GTX-2080 GPU) running Yolo-v5~\cite{glenn_jocher_2022_6222936} object detector. The captured frames from both cameras are sent for AU processing on the edge-server
over a \emph{5G} network with an average frame uploading latency of \emph{39.7}
ms. While the first
camera stream is sent just to the object detector AU, the second
camera stream is also sent to \approach\ which runs on a low-end IoT
device (Intel NUC box with a 2.6 GHz Intel i7-6770HQ CPU) in parallel
to object detector AU. We first perform in-situ training using the second
camera stream to populate the Q-table for SARSA RL agent for 12 hours
then we observe the performance of \approach\ for next consecutive
days in the exploitation phase. In this setup, \approach\ adjusts the camera
parameters every 30 seconds.

\begin{figure}[tb]
\begin{subfigure}[]{0.99\linewidth}
\centering
    \includegraphics[height=1.5in]{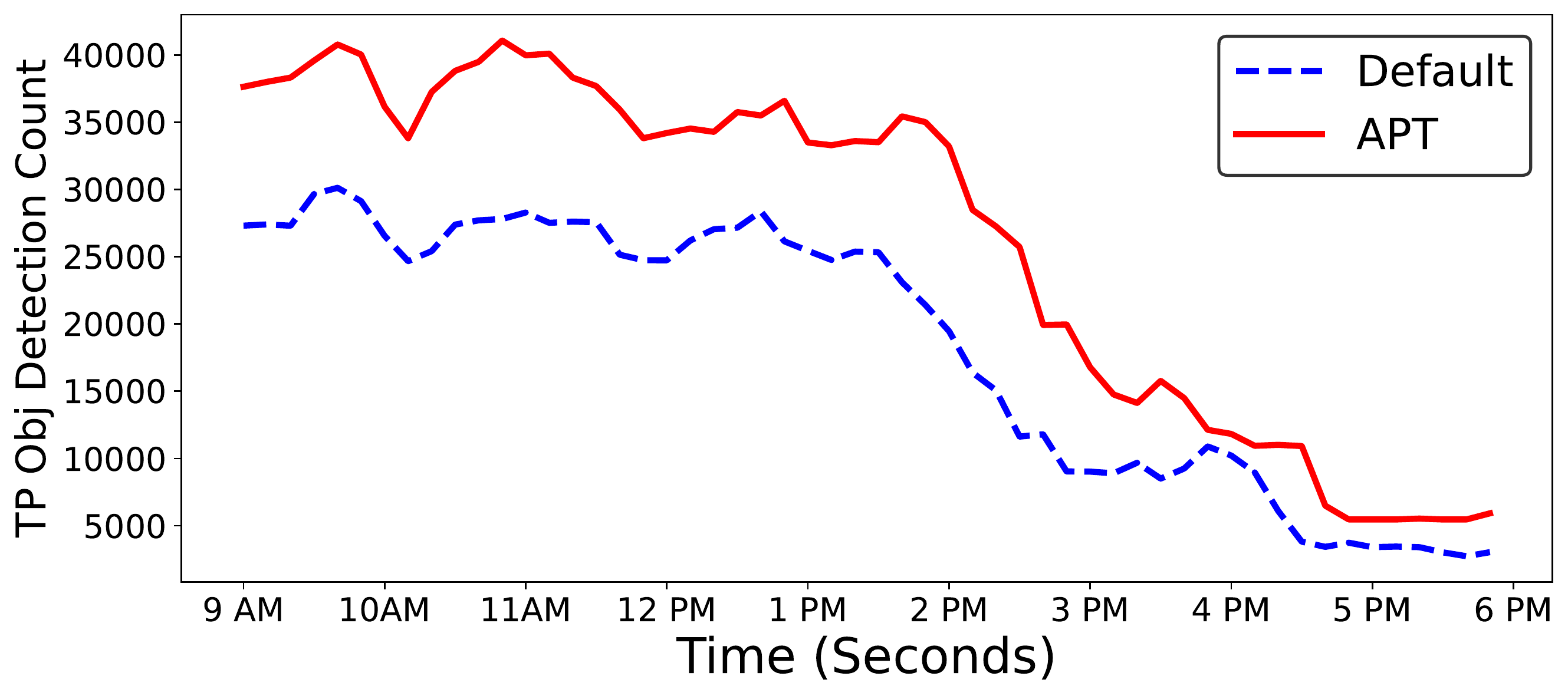}
    \caption{Day 1}
    \label{fig:daylong_real_deploy_1}
\end{subfigure}%
\hfill
\begin{subfigure}[]{0.99\linewidth}
\centering
    \includegraphics[height=1.5 in]{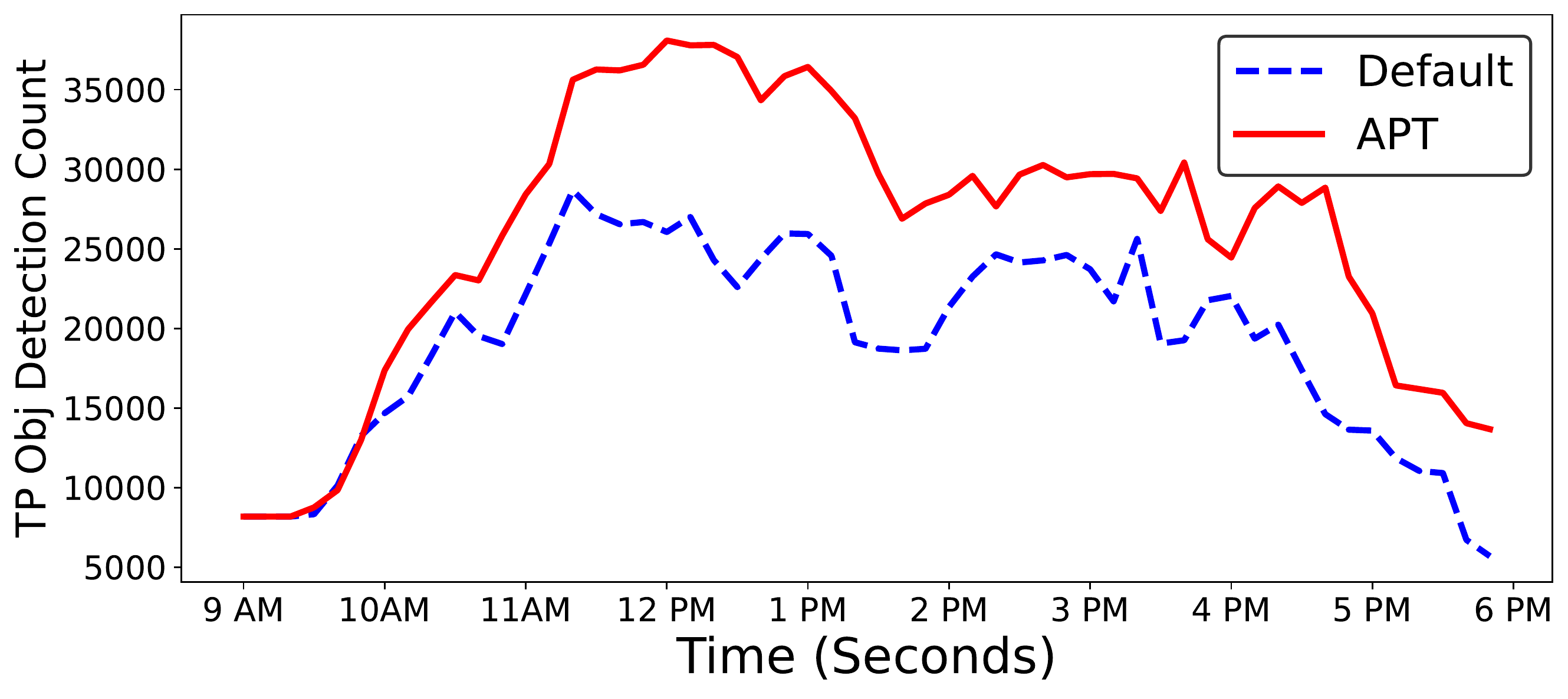}
    \caption{Day 2}
    \label{fig:daylong_real_deploy_2}
\end{subfigure}

\hfill

\begin{subfigure}[]{0.99\linewidth}
\centering
    \includegraphics[height=1.5 in]{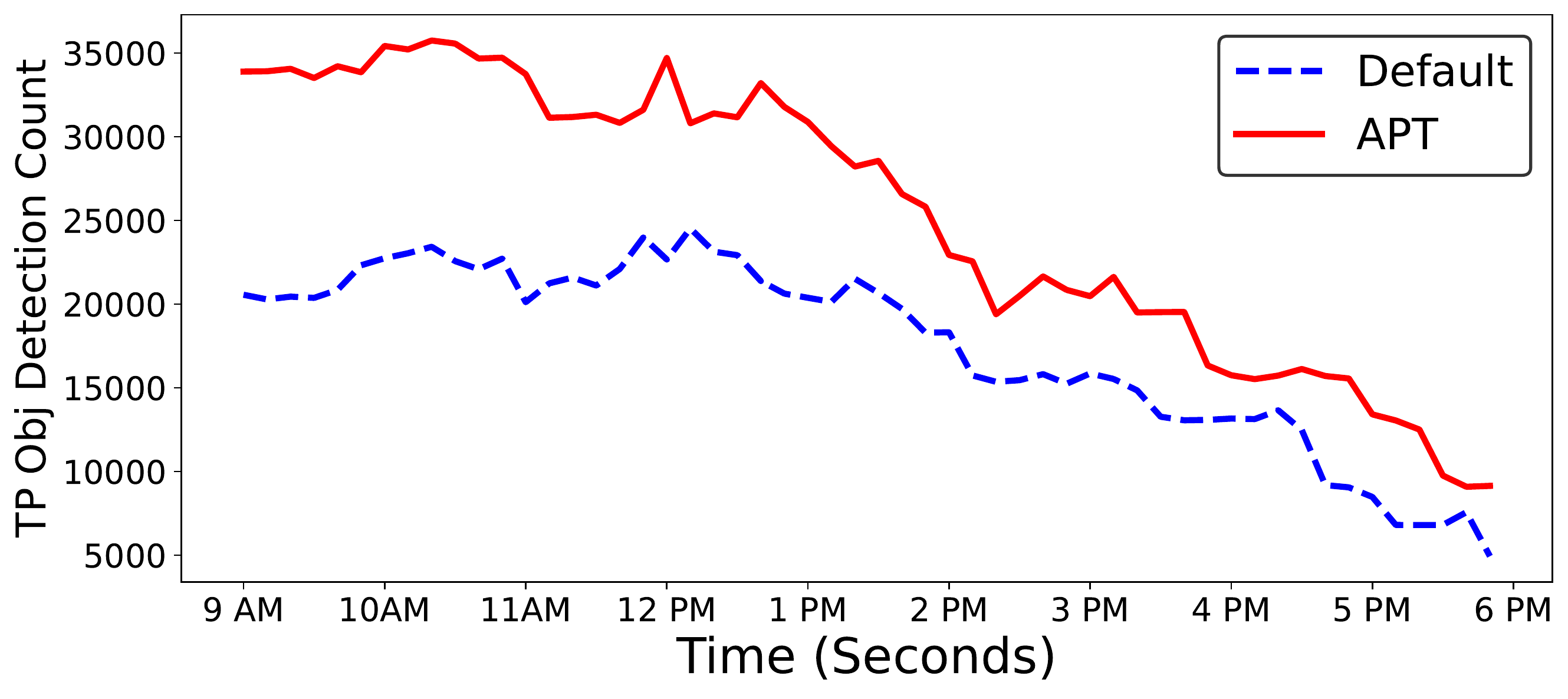}
    \caption{Day 3}
    \label{fig:daylong_real_deploy_3}
\end{subfigure}
 \caption{\approach performance (with Rank-IQA reward function) throughout the day in Parking lot (true-positive object detections are accumulated for each 10-minute interval).}
 \vspace{-0.2in}
  \label{fig:daylong_real}
\end{figure}

We ran both video analytics pipeline (VAP) for 9 continuous peak hours
in each day, i.e., during daylight. We also recorded the videos captured by the cameras and the
detections on those camera stream to manually inspect and validate the
detections from both VAPs. ~\figref{fig:daylong_real} shows the total
true-positive object detections (\ie car and person) in each 10-minute
 interval from 9AM-6PM for three consecutive days. 
% While, the analytical performance improvement observed by \approach\ over the default camera video stream varies throughout the day, 
\approach\ constantly provides higher
true-positive detection count than the default camera stream during all segments of the day, as shown in~\figref{fig:daylong_real}. We also
observe an improvement of \emph{44.71}\%, \emph{35.49}\% and
\emph{45.92}\% (average $\sim$ 42 \%) more true-positive object detections from the camera
stream tuned by \approach\ when compared to the default camera stream for first three
days of evaluation, respectively. Thus, \approach\ is effective in adaptively tuning camera parameters such that the video quality is improved, thereby resulting in improvement in video analytics accuracy.

%Thus, \approach\ can effectively adjust camera parameters based on the environment and content seen by the camera and helps in improving analytics performance.

\if 0
\begin{figure}[t]
    \centering
    \includegraphics[width=0.55\linewidth]{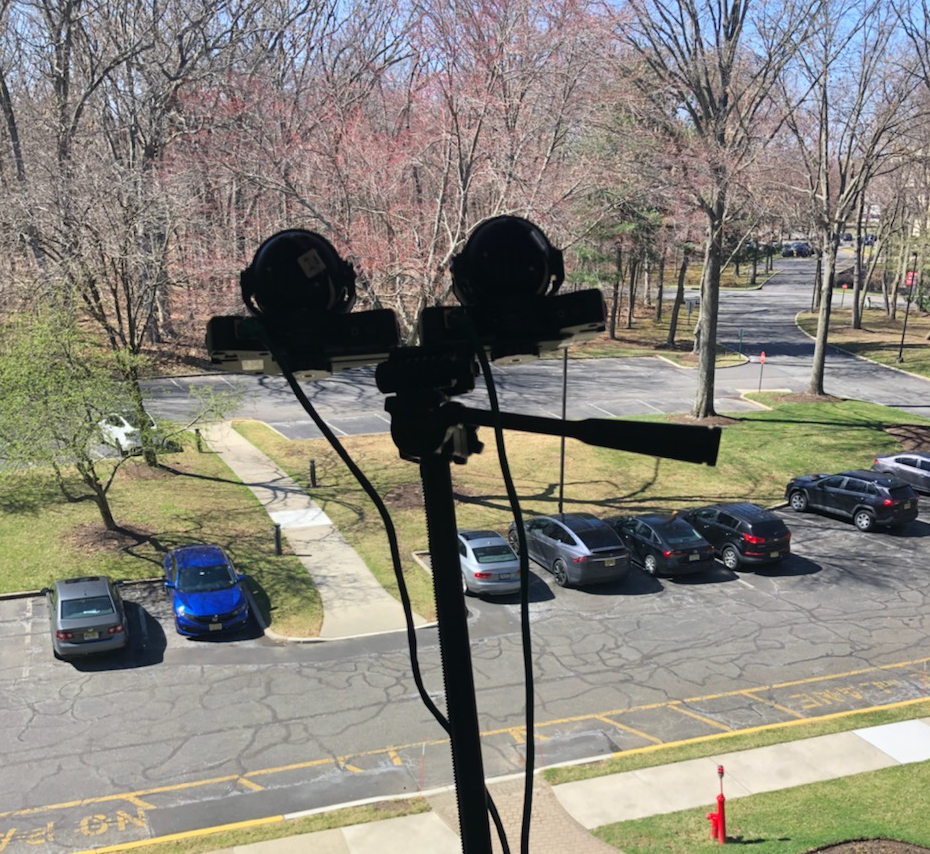}
    \caption{\approach real-world deployment setup.}
    \label{fig:real_setup}
%\vspace{-0.1in}
\end{figure}
\fi

\vspace{-0.05in}
\section{Future Works}
\label{sec:future}

Our demonstration in this paper that adaptive camera parameter tuning can improve
video analytics accuracy opens up several new research avenues. Here,
we designed \approach, which adaptively tunes four image-appearance
parameters based on the perceptual quality metric. In future
work, we plan to study how other non-automated image and
video-specific camera parameters such as max shutter speed, maximum
gain, compression, bitrate, FPS, \etc can be adpatively tuned to
enhance video analytics accuracy. In addition to video surveillance camera
sensor, we also plan to extend \approach\ design to tune the parameters of
other complex sensors such as depth and thermal cameras.

For \approach\ design, we borrowed the quality estimator from SOTA
perceptual no-reference IQA (\ie Rank-IQA). Since the captured content
are consumed by downstream AUs, in future work we plan to explore the
scope of quality estimation based on the downstream AU's perception,
similar to the work presented in~\cite{paul2021aqua}.

\vspace{-0.15in}
\section{Conclusion}
\label{sec:conc}

Video analytics applications heavily rely on good quality of video
input to produce accurate analytics results. In this paper,
we show that variation in environmental conditions and video content
can lead to degradation in input video quality, leading to degradation
of overall analytics insights. To mitigate this loss in accuracy, we
propose \approach, which uses reinforcement learning techniques to
adaptively tune camera parameters so as to improve video quality,
thereby improving accuracy of analytics. Through real-world
experiments, we show that \approach consistently performs
better than the fixed manufacturer-provided default camera settings,
%where camera parameters are fixed to the manufacturer-provided default,
and on average improves the accuracy of object detection video
analytics application by $\sim$ 42\%.

%\vspace{-0.1in}

\textbf{\textit{Acknowledgment.}}
This  project is supported in part 
by NEC Labs America
and by NSF grant 2211459-CNS.

\bibliographystyle{abbrv}
\bibliography{egbib}

\begin{thebibliography}{10}

\bibitem{pil}
Pillow library.
\newblock \url{https://pillow.readthedocs.io/en/stable/}.

\bibitem{bezryadin2007brightness}
S.~Bezryadin, P.~Bourov, and D.~Ilinih.
\newblock Brightness calculation in digital image processing.
\newblock In {\em International symposium on technologies for digital photo
  fulfillment}, volume 2007, pages 10--15. Society for Imaging Science and
  Technology, 2007.

\bibitem{Ekya}
R.~Bhardwaj, Z.~Xia, G.~Ananthanarayanan, J.~Jiang, Y.~Shu, N.~Karianakis,
  K.~Hsieh, P.~Bahl, and I.~Stoica.
\newblock Ekya: Continuous learning of video analytics models on edge compute
  servers.
\newblock In {\em 19th USENIX Symposium on Networked Systems Design and
  Implementation (NSDI 22)}, pages 119--135, Renton, WA, Apr. 2022. USENIX
  Association.

\bibitem{vapix}
A.~{C}ommunications.
\newblock Vapix library.

\bibitem{de2013sharpness}
K.~De and V.~Masilamani.
\newblock Image sharpness measure for blurred images in frequency domain.
\newblock {\em Procedia Engineering}, 64:149--158, 2013.

\bibitem{diamond2021dirty}
S.~Diamond, V.~Sitzmann, F.~Julca-Aguilar, S.~Boyd, G.~Wetzstein, and F.~Heide.
\newblock Dirty pixels: Towards end-to-end image processing and perception.
\newblock {\em ACM Transactions on Graphics (TOG)}, 40(3):1--15, 2021.

\bibitem{allied-market-research}
V.~Gaikwad and R.~Rake.
\newblock Video {A}nalytics {M}arket {S}tatistics: 2027, 2021.

\bibitem{hasler2003colormeasuring}
D.~Hasler and S.~E. Suesstrunk.
\newblock Measuring colorfulness in natural images.
\newblock In {\em Human vision and electronic imaging VIII}, volume 5007, pages
  87--95. International Society for Optics and Photonics, 2003.

\bibitem{heide2014flexisp}
F.~Heide, M.~Steinberger, Y.-T. Tsai, M.~Rouf, D.~Pajak, D.~Reddy, O.~Gallo,
  J.~Liu, W.~Heidrich, K.~Egiazarian, et~al.
\newblock Flexisp: A flexible camera image processing framework.
\newblock {\em ACM TOG}, 33(6):1--13, 2014.

\bibitem{jang2018application}
S.~Y. Jang, Y.~Lee, B.~Shin, and D.~Lee.
\newblock Application-aware {I}o{T} camera virtualization for video analytics
  edge computing.
\newblock In {\em IEEE/ACM SEC}, pages 132--144. IEEE, 2018.

\bibitem{jiang2018chameleon}
J.~Jiang, G.~Ananthanarayanan, P.~Bodik, S.~Sen, and I.~Stoica.
\newblock Chameleon: scalable adaptation of video analytics.
\newblock In {\em Proc. of ACM SIGCOMM}, pages 253--266, 2018.

\bibitem{glenn_jocher_2022_6222936}
G.~Jocher, A.~Chaurasia, A.~Stoken, J.~Borovec, NanoCode012, Y.~Kwon, TaoXie,
  J.~Fang, imyhxy, K.~Michael, Lorna, A.~V, D.~Montes, J.~Nadar, Laughing,
  tkianai, yxNONG, P.~Skalski, Z.~Wang, A.~Hogan, C.~Fati, L.~Mammana,
  AlexWang1900, D.~Patel, D.~Yiwei, F.~You, J.~Hajek, L.~Diaconu, and M.~T.
  Minh.
\newblock {ultralytics/yolov5: v6.1 - TensorRT, TensorFlow Edge TPU and
  OpenVINO Export and Inference}, Feb. 2022.

\bibitem{kang2014convolutional}
L.~Kang, P.~Ye, Y.~Li, and D.~Doermann.
\newblock Convolutional neural networks for no-reference image quality
  assessment.
\newblock In {\em Proceedings of the IEEE CVPR}, pages 1733--1740, 2014.

\bibitem{AMS}
M.~Khani, P.~Hamadanian, A.~Nasr-Esfahany, and M.~Alizadeh.
\newblock Real-time video inference on edge devices via adaptive model
  streaming.
\newblock In {\em 2021 IEEE/CVF International Conference on Computer Vision
  (ICCV)}, pages 4552--4562, 2021.

\bibitem{Kuehne11}
H.~Kuehne, H.~Jhuang, E.~Garrote, T.~Poggio, and T.~Serre.
\newblock {HMDB}: a large video database for human motion recognition.
\newblock In {\em Proceedings of ICCV}, 2011.

\bibitem{liu2020joint}
L.~Liu, X.~Jia, J.~Liu, and Q.~Tian.
\newblock Joint demosaicing and denoising with self guidance.
\newblock In {\em Proceedings of the IEEE/CVF CVPR}, pages 2240--2249, 2020.

\bibitem{liu2017rankiqa}
X.~Liu, J.~Van De~Weijer, and A.~D. Bagdanov.
\newblock Rankiqa: Learning from rankings for no-reference image quality
  assessment.
\newblock In {\em Proceedings of the IEEE ICCV}, pages 1040--1049, 2017.

\bibitem{brisque_mittal2012no}
A.~Mittal, A.~K. Moorthy, and A.~C. Bovik.
\newblock No-reference image quality assessment in the spatial domain.
\newblock {\em IEEE Transactions on image processing}, 21(12):4695--4708, 2012.

\bibitem{biqi_moorthy2010two}
A.~K. Moorthy and A.~C. Bovik.
\newblock A two-step framework for constructing blind image quality indices.
\newblock {\em IEEE Signal processing letters}, 17(5):513--516, 2010.

\bibitem{niebles2010modeling}
J.~C. Niebles, C.-W. Chen, and L.~Fei-Fei.
\newblock Modeling temporal structure of decomposable motion segments for
  activity classification.
\newblock In {\em ECCV}, pages 392--405. Springer, 2010.

\bibitem{nishimura2018automatic}
J.~Nishimura, T.~Gerasimow, R.~Sushma, A.~Sutic, C.-T. Wu, and G.~Michael.
\newblock Automatic isp image quality tuning using nonlinear optimization.
\newblock In {\em Proc. of IEEE ICIP}, pages 2471--2475. IEEE, 2018.

\bibitem{cvat}
openvinotoolkit.
\newblock Computer vision annotation tool (cvat).
\newblock \url{https://github.com/openvinotoolkit/cvat}.

\bibitem{NIST}
M.~N. Patrick~Grother and K.~Hanaoka.
\newblock Face {R}ecognition {V}endor {T}est ({FRVT}).
\newblock \url{https://nvlpubs.nist.gov/nistpubs/ir/2019/NIST.IR.8271.pdf},
  2019.

\bibitem{paul2021aqua}
S.~Paul, U.~Drolia, Y.~C. Hu, and S.~T. Chakradhar.
\newblock Aqua: Analytical quality assessment for optimizing video analytics
  systems.
\newblock In {\em IEEE/ACM SEC}, pages 135--147. IEEE, 2021.

\bibitem{video-analytics-accuracy-fluctuation}
S.~Paul, K.~Rao, G.~Coviello, M.~Sankaradas, O.~Po, Y.~C. Hu, and
  S.~Chakradhar.
\newblock Why is the video analytics accuracy fluctuating, and what can we do
  about it?
\newblock {\em arXiv preprint arXiv:2208.12644v2}, 2022.

\bibitem{peli1990contrast}
E.~Peli.
\newblock Contrast in complex images.
\newblock {\em JOSA A}, 7(10):2032--2040, 1990.

\bibitem{report-linker}
D.~Research.
\newblock Global {S}urveillance {C}amera {M}arket: {A}nalysis {B}y {S}ystem
  {T}ype, {B}y {T}echnology {B}y {R}egion {S}ize and {T}rends with {I}mpact of
  {COVID}-19 and {F}orecast up to 2027, 2022.

\bibitem{schwartz2018deepisp}
E.~Schwartz, R.~Giryes, and A.~M. Bronstein.
\newblock Deepisp: Toward learning an end-to-end image processing pipeline.
\newblock {\em IEEE Transactions on Image Processing}, 28(2):912--923, 2018.

\bibitem{Su_2020_CVPR}
S.~Su, Q.~Yan, Y.~Zhu, C.~Zhang, X.~Ge, J.~Sun, and Y.~Zhang.
\newblock Blindly assess image quality in the wild guided by a self-adaptive
  hyper network.
\newblock In {\em Proceedings of the IEEE/CVF CVPR}, June 2020.

\bibitem{introduction-to-rl}
R.~S. Sutton, A.~G. Barto, et~al.
\newblock {\em Introduction to reinforcement learning}, volume 135.
\newblock MIT press Cambridge, 1998.

\bibitem{q-learning}
C.~J. C.~H. Watkins and P.~Dayan.
\newblock Q-learning.
\newblock In {\em Machine Learning}, pages 279--292, 1992.

\bibitem{q-lambda}
M.~Wiering and J.~Schmidhuber.
\newblock Fast {O}nline q($\lambda$).
\newblock {\em Machine Learning}, 33(1):105--115, Oct 1998.

\bibitem{wu2019visionisp}
C.-T. Wu, L.~F. Isikdogan, S.~Rao, B.~Nayak, T.~Gerasimow, A.~Sutic,
  L.~Ain-kedem, and G.~Michael.
\newblock Visionisp: Repurposing the image signal processor for computer vision
  applications.
\newblock In {\em IEEE ICIP}, pages 4624--4628. IEEE, 2019.

\bibitem{gmlog_xue2014blind}
W.~Xue, X.~Mou, L.~Zhang, A.~C. Bovik, and X.~Feng.
\newblock Blind image quality assessment using joint statistics of gradient
  magnitude and laplacian features.
\newblock {\em IEEE Transactions on Image Processing}, 23(11):4850--4862, 2014.

\bibitem{zhang2018awstream}
B.~Zhang, X.~Jin, S.~Ratnasamy, J.~Wawrzynek, and E.~A. Lee.
\newblock Awstream: Adaptive wide-area streaming analytics.
\newblock In {\em Proceedings of the 2018 Conference of the ACM Special
  Interest Group on Data Communication}, pages 236--252, 2018.

\bibitem{201465}
H.~Zhang, G.~Ananthanarayanan, P.~Bodik, M.~Philipose, P.~Bahl, and M.~J.
  Freedman.
\newblock Live video analytics at scale with approximation and delay-tolerance.
\newblock In {\em Proc. of 14th {USENIX} NSDI)}, pages 377--392, Boston, MA,
  Mar. 2017. {USENIX} Association.

\bibitem{zhang2018ffdnet}
K.~Zhang, W.~Zuo, and L.~Zhang.
\newblock Ffdnet: Toward a fast and flexible solution for cnn-based image
  denoising.
\newblock {\em IEEE Transactions on Image Processing}, 27(9):4608--4622, 2018.

\end{thebibliography}

\end{document}